\PassOptionsToPackage{table}{xcolor}
\documentclass[sigconf,anonymous=false]{acmart}

\usepackage{booktabs} % For formal tables
\usepackage{subcaption}
\setcopyright{acmcopyright}
\acmDOI{xx.xxx/xxx_x}

\acmISBN{979-8-4007-0629-5/25/03}

\acmConference[SAC'25]{ACM SAC Conference}{March 31 –April 4, 2025}{Sicily, Italy}
\acmYear{2025}
\copyrightyear{2025}

\acmArticle{4}
\acmPrice{15.00}

\begin{document}
\title{Efficient Pruning of Text-to-Image Models: Insights from Pruning Stable Diffusion}
% \titlenote{Produces the permission block, and
% \subtitle{Extended Abstract}
% \subtitlenote{The full version of the author's guide is available as
  % \texttt{acmart.pdf} document}
  
%\renewcommand{\shorttitle}{Efficient Pruning of Text-to-Image Models: A Comprehensive Study on Stable Diffusion}

\author{Samarth N Ramesh}
% \authornote{Dr.~Trovato insisted his name be first.}
\affiliation{%
  \institution{University of Sheffield}
  % \streetaddress{P.O. Box 1212}
  \city{Sheffield} 
  \country{United Kingdom}
}
\email{snramesh1@sheffield.ac.uk}

\author{Zhixue Zhao}
% \authornote{The secretary disavows any knowledge of this author's actions.}
\affiliation{%
  \institution{University of Sheffield}
  % \streetaddress{P.O. Box 1212}
  \city{Sheffield} 
  \country{United Kingdom}
}
\email{zhixue.zhao@sheffield.ac.uk}

% \renewcommand{\shortauthors}{B. Trovato et al.}

\begin{abstract}
%Deep learning models are achieving remarkable results in a wide range of fields, but these results come with a proportionate increase in model size. These growing model sizes create the need for model compression techniques. For deep learning models, the two most common model compression approaches are quantization and pruning. In this dissertation, we study one-shot pruning for text-to-image models, in particular Stable Diffusion 2. Stable Diffusion 2 consists of an image component and a text component, both of which have different sizes and very different structures. We explore how well existing one-shot pruning techniques for pruning text models and image models perform when applied to text-to-image models. We study how the model behaves when only one component is pruned at a time and find that the image quality drops off sharply at a threshold instead of linearly deteriorating. We propose an integrated approach for simultaneously pruning both components of the text-to-image model together and rigorously evaluate these pruned models using the industry standard FID and CLIP Score metrics. Finally, we propose a configuration of optimal pruning techniques and sparsities for Stable Diffusion 2.

As text-to-image models grow increasingly powerful and complex, their burgeoning size presents a significant obstacle to widespread adoption, especially on resource-constrained devices. This paper presents a pioneering study on post-training pruning of Stable Diffusion 2, addressing the critical need for model compression in text-to-image domain. 
Unlike previous work focused on language models or traditional image generation, our study tackles the complex multimodality generation models, and particularly examines the pruning impact on the textual component and the image generation component separately.
We conduct a comprehensive comparison on pruning the model or the single component of the model in various sparsities. Our results yield previously undocumented findings. For example, contrary to established trends in language model pruning, we discover that simple magnitude pruning outperforms more advanced techniques in text-to-image context. Furthermore, our results show that Stable Diffusion 2 can be pruned to 38.5\% sparsity with minimal quality loss, achieving a significant reduction in model size. We propose an optimal pruning configuration that prunes the text encoder to 47.5\% and the diffusion generator to 35\%. This configuration maintains image generation quality while substantially reducing computational requirements.
In addition, our work uncovers intriguing questions about information encoding in text-to-image models: we observe that pruning beyond certain thresholds leads to sudden performance drops (unreadable images), suggesting that specific weights encode critical semantics information. This finding opens new avenues for future research in model compression, interoperability, and bias identification in text-to-image models.
By providing crucial insights into the pruning behavior of text-to-image models, our study lays the groundwork for developing more efficient and accessible AI-driven image generation systems.\footnote{We publicly released our code at: \texttt{https://github.com/samarthramesh/SD2-Pruning}}

\end{abstract}

\begin{CCSXML}
<ccs2012>
   <concept>
       <concept_id>10010147.10010178.10010224</concept_id>
       <concept_desc>Computing methodologies~Computer vision</concept_desc>
       <concept_significance>500</concept_significance>
       </concept>
   <concept>
       <concept_id>10010147.10010178.10010179</concept_id>
       <concept_desc>Computing methodologies~Natural language processing</concept_desc>
       <concept_significance>500</concept_significance>
       </concept>
 </ccs2012>
\end{CCSXML}

\ccsdesc[500]{Computing methodologies~Computer vision}
\ccsdesc[500]{Computing methodologies~Natural language processing}

\keywords{Computer Vision, Natural Language Processing, Model Compression, Pruning, Text-to-Image Generation}

\maketitle

\section{Introduction}

Large language models (LLMs) and diffusion models often face the challenge of having exceptionally large model sizes~\cite{ko2023large, brown2020language}. While these extensive parameters allow them to understand high-quality text and generate images, they also result in substantial computational demands and resource consumption~\cite{effficient_large-scale, brown2020language, diffusion_model_comprehensive_survey}. With billions of parameters, these models become prohibitively expensive to operate, limiting their use primarily to large corporations. To address this issue, significant research has been devoted to model compression techniques for deep learning models. Compressing these models can make them deployable on edge devices~\cite{prakash2022iot, quantization_deployment_microcontrolers, dnn_deployment}, reduce response times~\cite{hawks2021ps, liang2021pruning}, enable real-time applications~\cite{niu2020patdnn, quantization_deployment_microcontrolers}, and greatly increase their accessibility.

% There were two dominant approaches to deep learning model compression, namely quantization and pruning. Given a model, quantization aims to reduce the precision of the model's weights while pruning attempts to identify and remove redundant weights. Due to the recent explosion in model size, quantization has enjoyed more popularity in practice due to its scalability~\cite{sheng2023flexgen, hoffmann2022training}. Pruning has comparatively received less attention in theory and practice due to the need for retraining which is time-consuming and expensive. However, more recently there has been research on post-training pruning methods for LLMs that can deliver a compressed model without the need for retraining~\cite{frantar_sparsegpt_2023, sun_simple_2023}. 

Quantization~\cite{gray_quantization_1998} and pruning~\cite{lecun_optimal_1989, hassibi1993optimal, han_learning_2015} are two primary compression approaches. Quantization focuses on reducing the precision of a model's weights, while pruning seeks to identify and eliminate redundant weights. %As model sizes have grown dramatically in recent years, quantization has gained wider adoption due to its scalability and practical feasibility~\cite{sheng2023flexgen, hoffmann2022training}. In contrast, pruning has received less attention both in theory and in practice, largely because it typically requires retraining, a process that is both time-consuming and computationally expensive. 
Both methods have been well studied for LLMs but much less has been done for vision language models~\citep{LOPES2024104015}. 
Further, recent advancements have introduced post-training pruning methods for LLMs, which offer compressed models without the need for retraining~\cite{frantar_sparsegpt_2023, sun_simple_2023}.% This work focuses on pruning text-to-image models for it is a particularly under-explored direction.

% There has comparatively been less work on post-training pruning for image diffusion models and to the author's knowledge, there has been no research on post-training pruning of text-to-image models. This paper \cass{could you read through the paper, to remove the dissertation trace, it is a paper} aims to bridge this gap by exploring the effectiveness of post-training pruning on Stable Diffusion 2, a text-to-image model by Stability AI. We study how individually pruning the text and the image components of the model affects the performance of the model and obtain optimal sparsities for each component of Stable Diffusion 2. Finally, we obtain an optimal sparsity configuration for pruning the entire Stable Diffusion 2 model.

However, there has been relatively little research on post-training pruning for vision language models. As far as we know, no prior work has addressed post-training pruning of text-to-image models. This paper seeks to fill this gap by investigating the impact of post-training pruning on Stable Diffusion 2~\citep{rombach_high-resolution_2022}. Additionally, this study examines how pruning impacts the textual encoder and the image generator of the model separately, and identifies the optimal sparsity levels for each component. %Additionally, the paper proposes an optimal sparsity configuration for pruning the entire Stable Diffusion 2 model.

\section{Related Work}

\subsection{Text-to-Image Models}

% \textcolor{blue}{cass: different t2i models, first talk about sth not diffusion, one short paragraph}

The field of image generation was initially dominated by Generative Adversarial Networks (GANs)~\cite{goodfellow_generative_2014, brock2018large} and Diffusion models~\cite{sohl-dickstein_deep_2015}. While GANs produce high-quality images, their variability is constrained by data used in the adversarial training process and they require substantial effort to scale for diverse, complex applications~\cite{karras2019style}. On the other hand, Diffusion models, built on denoising autoencoders, excel across various image synthesis tasks~\cite{ho_denoising_2020, song2020score}, but their high computational demands, requiring extensive GPU resources for training and inference, present significant challenges~\cite{dhariwal2021diffusion}.

% The Stable Diffusion models released by Stability AI have each brought about significant improvement in the field of text-to-image generation and are state-of-the-art models~\cite{rombach2021highresolution, esser_scaling_2024}. These models extend diffusion models by introducing a variational autoencoder that represents images in a latent space, which has many advantages over training diffusion models in the image space~\cite{rombach2021highresolution}. The diffusion models are applied in the latent space and cross-attention layers are used to condition on rich representation of the input prompt. In the case of text-to-image applications, this rich representation is provided by a transformer model~\cite{rombach_high-resolution_2022}. For the purpose of this paper, we chose to focus on Stable Diffusion 2 which uses the text encoder from the CLIP model~\cite{radford_learning_2021} as the transformer and a U-Net~\cite{ronneberger2015u} as the Diffusion model. Stable Diffusion 2 is state-of-the art and it's structure is a core building block for later models~\cite{}. Stable Diffusion 2 has 1.2 billion parameters of which 340 million parameters are in the CLIP text encoder and 860 million parameters are in the U-Net diffusion generator. Its large size makes it an ideal candidate for model compression techniques.

The Stable Diffusion models released by Stability AI have made significant advances in text-to-image generation and are recognized as state-of-the-art~\cite{rombach2021highresolution, esser_scaling_2024}. These models build upon diffusion models by incorporating a variational autoencoder to encode images in a latent space, offering several advantages over training in the image space, such as significantly reducing the computational resources needed for both training and inference~\cite{rombach2021highresolution}. The diffusion process is applied in this latent space, with cross-attention layers used to condition the model on rich representations of the input. For text-to-image tasks, this representation is generated by a transformer model~\cite{rombach_high-resolution_2022}. In this study, we focus on Stable Diffusion 2, which uses the CLIP model's text encoder~\cite{radford_learning_2021} as the transformer and a U-Net~\cite{ronneberger2015u} as the diffusion model. Stable Diffusion 2, a state-of-the-art model and a foundational architecture for later versions~\cite{esser_scaling_2024, ramesh2022hierarchical}, contains 1.2 billion parameters, with 340 million in the CLIP text encoder and 860 million in the U-Net diffusion generator. Its large size makes it an ideal candidate for model compression techniques.

% \textcolor{blue}{one paraphare for stable diffusion, and one sentence to say this paper mainly use stable diffusion as it is commonly use state-of-the-art}

\subsection{Pruning}

% \paragraph{Magnitude Pruning} The most naive approach is magnitude-based pruning \cite{han_learning_2015}, also known as magnitude pruning. Magnitude pruning works on the principle that the importance of a connection is directly proportional to the modulus of the weight value of the connection. Given by equation \ref{magnitude_pruning}, all weights are sorted and the portion of weights that are asserted to be of the least importance are pruned and their values are set to zero. While it does not provide the best performance, it is very simple and efficient to implement.
Pruning works on the principle that some connections in a network are of less importance than others~\cite{lecun_optimal_1989}. The method of finding these connections is of utmost importance for the post-training pruning. We introduce state-of-the-art pruning methods experimented in LLMs and their main difference is how they identify the weights to be pruned.

\paragraph{Magnitude Pruning}
Magnitude-based pruning~\cite{han_learning_2015} is one of the simplest and most commonly used techniques for pruning neural networks. It operates under the assumption that a connection's importance correlates with the absolute value of its weight. As described in Equation \ref{magnitude_pruning}, the weights are ranked according to their magnitude, and the least important weights are pruned by setting them to zero. While this method may not provide optimal performance compared to more advanced techniques, it is easy to implement, computationally efficient, and serves as a strong baseline for further refinement.

\begin{equation} \label{magnitude_pruning}
    \text{Low Importance Connections} = \operatorname*{argmin}_i |w_i|
\end{equation}

\paragraph{SparseGPT}

% For LLMs, several pruning methods have been developed that provide better post-training pruning performance than magnitude pruning. One of these was SparseGPT \cite{frantar_sparsegpt_2023} which attempts to solve an approximate version of the problem of finding the optimal pruning mask for a given set of parameters. This is simply a problem of finding a sparse subset of weights that provide the least reconstruction error. This is a problem that takes $O(d^4)$ time where $d$ is the dimension of the hidden layer in the model. SparseGPT arrives at an approximate solution in $O(d^3)$. This method outperformed magnitude pruning on LLMs of all sizes. 
% SparseGPT~\cite{frantar_sparsegpt_2023} tackles an approximate version of the problem of identifying the optimal pruning mask for a given set of parameters. 

SparseGPT~\cite{frantar_sparsegpt_2023} tackles the task by attempting to solve the problem of identifying the optimal pruning mask. The goal is to find a sparse subset of weights that minimizes reconstruction error. While the exact solution to this problem requires $O(d^4)$ time complexity~\cite{frantar_sparsegpt_2023}, where $d$ is the dimension of the hidden layer, SparseGPT achieves an approximate solution in $O(d^3)$, offering a more efficient approach. This method has consistently outperformed magnitude pruning across LLMs of various sizes.

% A more recent method that performs that improves upon SparseGPT and is used in many text-based models is Weight and Activation Pruning (Wanda) \cite{sun_simple_2023}. Wanda is motivated by the observation that emergent properties in LLMs are correlated with the presence of ``outliers''. These outliers are extremely large output values from neurons that are about 100 times larger than typical output values. These activations are considered ``outliers'' and are a good metric of the importance of certain weights. Wanda extends magnitude pruning by also considering the magnitude of the activations that are provided as input for these weights. The scoring metric is therefore provided by Equation \ref{wanda_scoring} where $W_{ij}$ is the value of the weights, $X_j$ are the activation provided as input to $W_{ij}$, and $||X_j||_2$ is the $l_2$ norm.

% \paragraph{Wanda} A more recent method that improves upon the performance of SparseGPT and is widely utilized in various text-based models is Weight and Activation Pruning (Wanda)~\cite{sun_simple_2023}. 

\paragraph{Wanda} Weight and Activation Pruning (Wanda)~\cite{sun_simple_2023} is a more recent method that surpasses SparseGPT in performance and has become widely adopted in various text-based models. Wanda is grounded in the observation that emergent properties in LLMs are often associated with the presence of ``outliers''~\cite{dettmers2022gpt3}. These outliers are defined as exceptionally large output values from neurons, which can be up to 100 times greater than typical output values. Such activations serve as valuable indicators of the importance of specific weights. 

Wanda extends the magnitude pruning by incorporating the values of the activations that serve as inputs for these weights. The scoring metric for this method is expressed in Equation \ref{wanda_scoring}, where \(W_{ij}\) represents the value of the weights, \(X_j\) denotes the activations input to \(W_{ij}\), and \(\|X_j\|_2\) is the \(l_2\) norm of these activations. This approach allows for a more nuanced assessment of weight importance, contributing to improved pruning effectiveness.

\begin{equation} \label{wanda_scoring}
    S_ij = |W_{ij}| . ||X_j||_2
\end{equation}

\paragraph{Outlier-Weighted Layerwise Pruning (OWL)}
% Another important observation on outliers in large language models is that the density of observed outliers in a particular layer of a model represents the importance of that particular layer to emergent properties. Based on this work, Outlier Weighted Layerwise pruning (OWL) \cite{yin_outlier_2024} aims to distribute the sparsity into the model's different layers in an unequal fashion. The required layer-wise sparsity is weighted by the number of outliers present, so more important layers are pruned less, while less important ones are pruned more. OWL is used in conjunction with other pruning algorithms such as Magnitude and Wanda, as it does not provide a notion of which weights to prune rather it provides recommended local sparsities. 

A significant observation regarding outliers in LLMs is that the density of observed outliers within a particular layer correlates with that layer's importance to the model's emergent properties~\cite{kovaleva2021bert, puccetti2022outliers, timkey2021all}. Based on this insight, Outlier-Weighted Layerwise Pruning (OWL)~\cite{yin_outlier_2024} seeks to distribute sparsity across different layers of the model in an uneven manner. The required layer-wise sparsity is influenced by the number of outliers present; consequently, layers deemed more important are pruned less aggressively, while those considered less critical are pruned more extensively. OWL can be integrated with other pruning algorithms, such as Magnitude and Wanda as it does not specify which weights to prune; rather, it provides recommendations for local sparsity levels within the layers.

While post-training pruning methods have been extensively studied for language models, their effectiveness on text-to-image models remains largely unexplored. Given the distinct architectures and training protocols of text-to-image models, it is challenging to infer how these pruning techniques, originally developed for language models, will perform in this different context.

\section{Experimental Setup}

This paper investigates the pruning of Stable Diffusion 2, a widely used text-to-image generation model. Stable Diffusion 2 has 1.2 billion parameters of which 340 million parameters (28\%) are in the CLIP text encoder and 860 million parameters (72\%) are in the U-Net diffusion generator. Our initial experiments focus on examining the effects of independently pruning individual components of the model while maintaining the integrity of the other parts. For example, only prune the text encoder component and keep the diffusion generator untouched. Following this, full model pruning is explored across multiple axes to determine the optimal balance of sparsities between the text encoder and the diffusion generator. The evaluation of the best configuration for full model pruning is informed by the results obtained from the individual component pruning experiments. %All experimentation is quantitatively evaluated using FID~\cite{heusel_gans_2018} and CLIP Score~\cite{hessel_clipscore_2022} metrics on 10,000 images generated from captions present in the MSCOCO 2017 dataset~\cite{lin_microsoft_2015}. These metrics are reinforced by qualitative evaluation of generated image quality.

% This paper is to prune Stable Diffusion 2, a well-established baseline in the field of text-to-image generation. The initial experiments were performed with the focus of studying the effect of pruning individual components of the model separately while keeping the rest of the model unchanged. Then the full model pruning was explored across multiple axes so that recommendations could be made for the best ratio between text and image sparsities. The testing for the optimal full model pruning configuration was guided by the results of the individual component pruning section.

% CLIP Text Encoder Pruning and U-Net Diffusion Generator Pruning
\subsection{Pruning Single Component}\label{sec:pruning_single_component}

% For text-encoder-only pruning, four different techniques were tested, namely magnitude pruning, Wanda, magnitude pruning with OWL and Wanda with OWL. For each of these techniques, multiple sparsities were tested in steps of 10\% and more granular tests were performed at intervals of interest. During initial experimentation and code implementation, most of the evaluation was based on a qualitative assessment of the generated images, but later these were reinforced with quantitative metrics.

For text-encoder-only pruning, four different techniques are tested, namely magnitude pruning, Wanda, magnitude pruning with OWL and Wanda with OWL. For each of these techniques, multiple sparsities are tested in steps of 10\% and more granular tests are performed at intervals of interest.

% So far there has been minimal work in post-training pruning for diffusion models. Some research has found efficient pruning techniques with retraining \cite{fang_structural_2023} and there was recently a proposal of a new technique specifically for iteratively pruning Latent Diffusion models \cite{castells_ld-pruner_2024} such as Stable Diffusion. However, the scope of the paper is limited to only magnitude pruning for the diffusion portion of the model, leaving other techniques to future work. Once again, multiple sparsities were tested in steps of 10\% and more granular tests were performed at intervals of interest.

To date, there has been limited research on post-training pruning specifically for diffusion models. Some studies have identified efficient pruning techniques that involve retraining~\cite{fang_structural_2023}, and a recent proposal introduced a novel approach for iteratively pruning Latent Diffusion models~\cite{castells_ld-pruner_2024}, such as Stable Diffusion. However, the scope of this paper is limited to magnitude pruning for the diffusion component of the model, leaving the exploration of other techniques for future research. As with the text encoder, multiple sparsity levels are evaluated in increments of 10\%, and more detailed assessments are conducted at intervals of particular interest.

\subsection{Full Model Pruning}\label{sec:pruning_full}

Two distinct approaches are employed for the experiments to find the optimal configuration for full model pruning. The first approach aims to determine the best ratio of sparsities between the text encoder and image diffusion generator if a specific full model sparsity is desired. At various levels of full model sparsity, the distribution of this sparsity between the two components is outlined in Table \ref{tab:full_model_sparsities_table}. It is important to note that certain ratios associated with higher full model sparsities are unfeasible due to the significantly smaller size of the text encoder compared to the diffusion generator. For example, when aiming to prune the full model to 50\%, attempting to distribute the sparsity in a 75:25 ratio for text and image would result in pruning the text encoder to 177\%, which is clearly not achievable.

\begin{table}
    \centering
    \scalebox{0.85}{
    \begin{tabular}{cccc}  
        \hline
        Full Model Sparsity & Text:Image Ratio & Text Sparsity & Image Sparsity \\
        \hline
        20\% & 75:25 & 53\% & 7\% \\
        20\% & 50:50 & 35\% & 14\% \\
        20\% & 25:75 & 18\% & 21\% \\
        \hline
        30\% & 75:25 & 80\% & 10\% \\
        30\% & 50:50 & 53\% & 21\% \\
        30\% & 25:75 & 27\% & 31\% \\
        \hline
        40\% & 50:50 & 71\% & 28\% \\
        40\% & 25:75 & 35\% & 42\% \\
        \hline
        50\% & 50:50 & 89\% & 35\% \\
        50\% & 25:75 & 44\% & 52\% \\
        \hline
        60\% & 25:75 & 53\% & 63\% \\
        \hline
    \end{tabular}
    }
    \caption{Sparsity is distributed between components in the ratios shown in Text:Image Ratio and Text and Image Sparsities represent the fraction of weights pruned in those components}
    \label{tab:full_model_sparsities_table}
\end{table}

The second approach to full model pruning relies on the findings from the individual component pruning experiments. Both the text encoder and the diffusion generator exhibit identifiable drop-off points in performance, which will be further examined in the results section. Assuming that the sub-models remain largely unaffected until reaching these drop-off points, the full model is pruned to align with the drop-off thresholds of both sub-models. This leads to a recommended configuration for maximal pruning that balances performance and sparsity effectively.

\subsection{Dataset}\label{sec:dataset}

We use the Microsoft COCO: Common Objects in Context (MSCOCO) 2017 dataset \cite{lin_microsoft_2015} which has a large number of real images and corresponding captions. MSCOCO is a commonly used dataset for computer vision tasks. For our experiments, we use 10,000 randomly sampled images and their corresponding captions.  

\subsection{Evaluation}\label{sec:evaluation}

\paragraph{FID}

The Fréchet Inception Distance (FID) \cite{heusel_gans_2018} defines a distance metric between two sets of images and measures how different the two sets are. A lower FID indicates that the two sets of images are very similar.

In our experiments, we use the FID to compare images generated from MSCOCO captions with corresponding real images. For each model we generate 10,000 images to calculate the FID.

\paragraph{CLIP Score}

The CLIP Score \cite{hessel_clipscore_2022} uses a multi-modal text and image model to directly compare a generated image with the provided prompt. It uses the CLIP model to calculate the similarity and a lower CLIP Score indicates a higher correlation in the semantic content of the prompt and image.

In our experiments, for each pruned model, we generate 10,000 images and calculate the average similarity of each image with its prompt using the CLIP Score.

\section{Results and Analysis}

% We aim to analyze the pruning impact on the text-to-image models, particularly we analyze the different components of stable diffusion separately, namely the CLIP text encoder and U-net diffusion generator, aiming to explore the optimal trade-off between performance and computational resources of the model. We first approached this problem by studying how pruning affects the model when pruning only one component at a time. We then use these insights to inform how we approach full model pruning.

The primary objective of this study is to analyze the impact of pruning on text-to-image models, specifically focusing on the individual components of Stable Diffusion 2, namely the CLIP text encoder and the U-Net diffusion generator. Our aim is to explore the optimal trade-off between model performance and computational resource efficiency. We initially approach this problem by examining the effects of pruning when targeting only one component at a time. The insights gained from this analysis subsequently inform our strategy for full model pruning.

All experimentation is quantitatively evaluated using FID~\cite{heusel_gans_2018} and CLIP Score~\cite{hessel_clipscore_2022} metrics on 10,000 images generated from captions present in the MSCOCO 2017 dataset~\cite{lin_microsoft_2015}. These metrics are reinforced by qualitative evaluation of generated image quality.

\subsection{CLIP Pruning Only}

% We experiment with two pruning methods on pruning the CLIP text encoder only. In addition to these, we also experiment with applying OWL onto these techniques to test whether outlier weighting provides similar improvement on text-to-image models. Figures~\ref{fig:sd2_text_magnitude_only_plots, fig:sd2_text_wanda_only_plots, fig:sd2_text_owl_wanda_only_plots, fig:sd2_text_owl_magnitude_only_plots} show FID and CLIP Scores for each of these methods. Pruning at low sparsities incur minimal losses across all methods. As sparsity increases, the performance of the model observes a sharp drop-off at a threshold in each method. Qualitative examination of the generated images support these observations. 

\begin{figure}
    \centering
    \includegraphics[width=\linewidth]{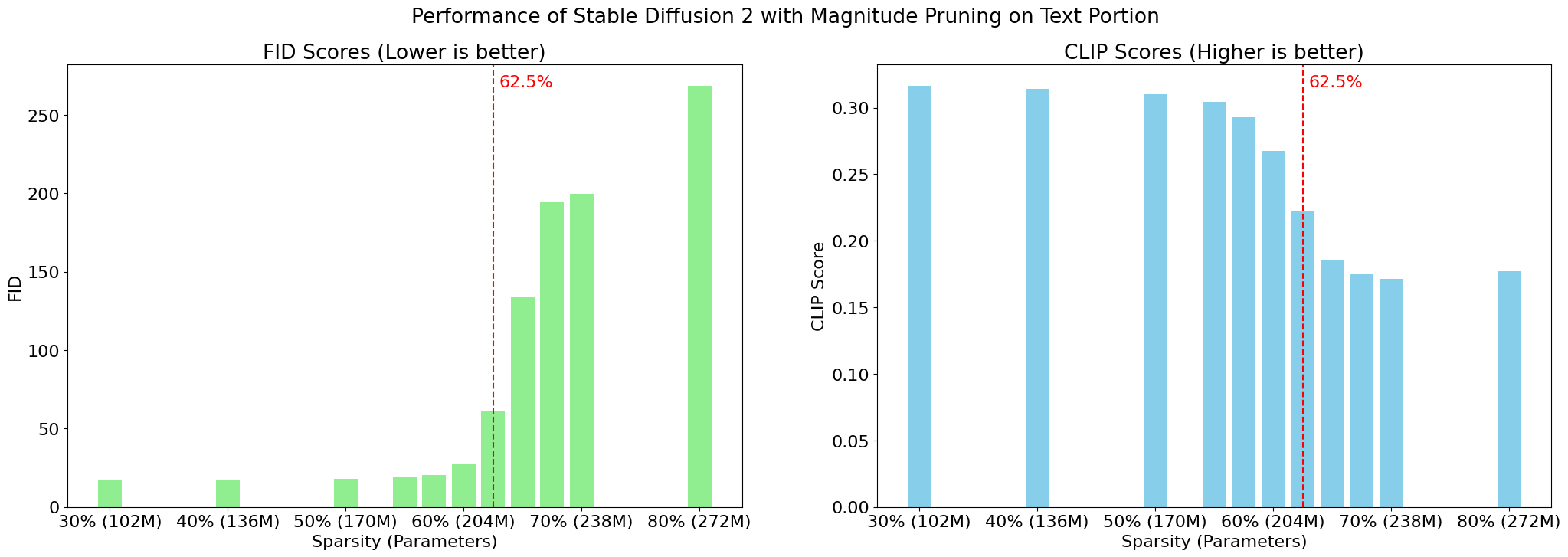}
    \caption{Pruning only Text Encoder of Stable Diffusion 2 using Magnitude Pruning - Sharp drop off in performance observed at 62.5\%}
    \label{fig:sd2_text_magnitude_only_plots}
\end{figure}

\begin{figure}
    \centering
    \includegraphics[width=\linewidth]{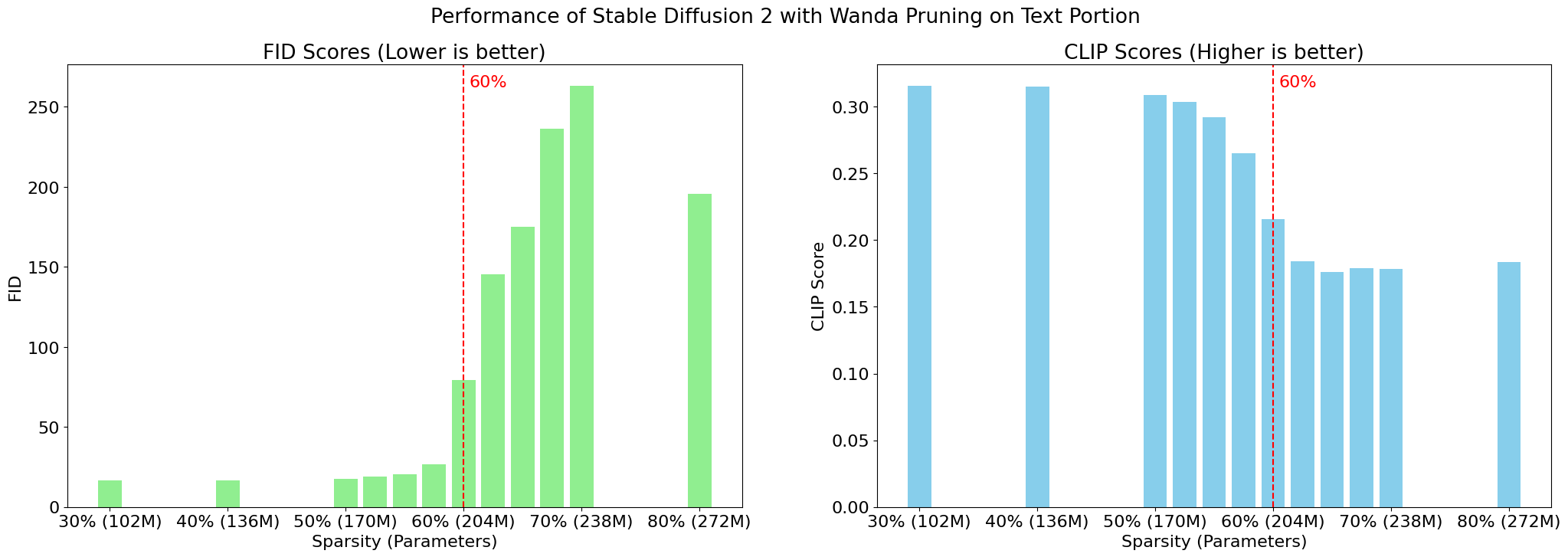}
    \caption{Pruning only Text Encoder of Stable Diffusion 2 using Wanda Pruning - Sharp drop off in performance observed at 60\%}
    \label{fig:sd2_text_wanda_only_plots}
\end{figure}

\begin{figure}
    \centering
    \includegraphics[width=\linewidth]{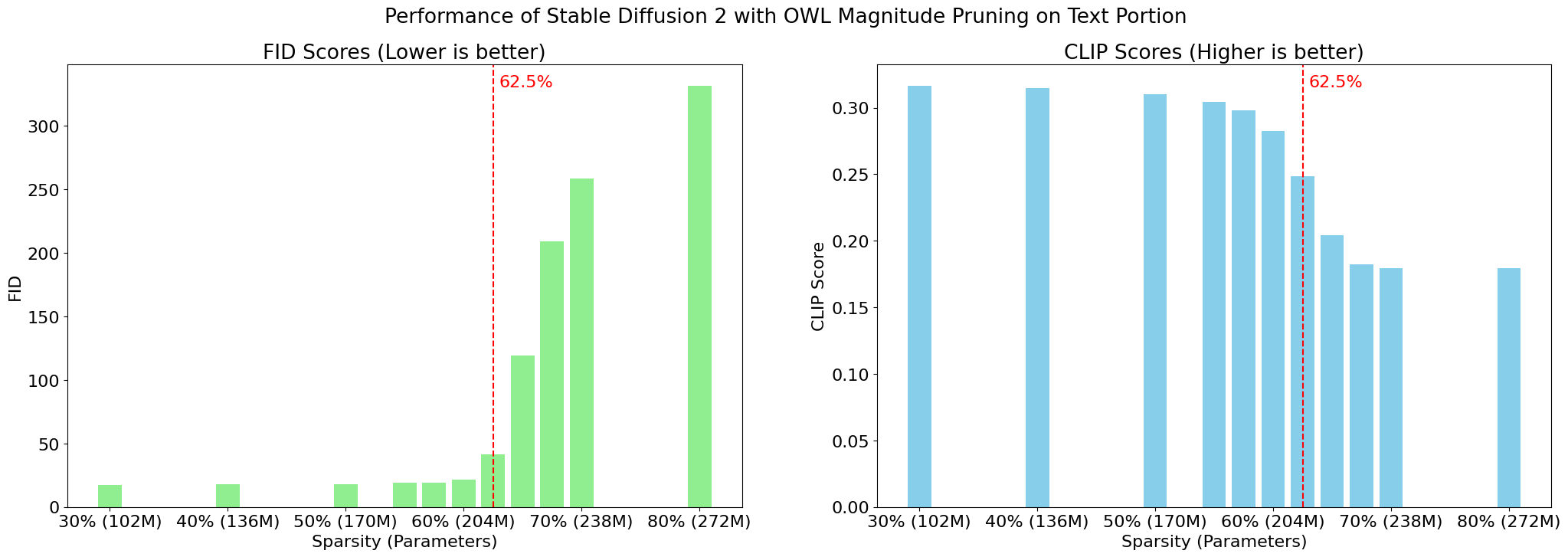}
    \caption{Pruning only Text Encoder of Stable Diffusion 2 using Magnitude Pruning with OWL - Sharp drop off in performance observed at 62.5\%}
    \label{fig:sd2_text_owl_magnitude_only_plots}
\end{figure}

\begin{figure}
    \centering
    \includegraphics[width=\linewidth]{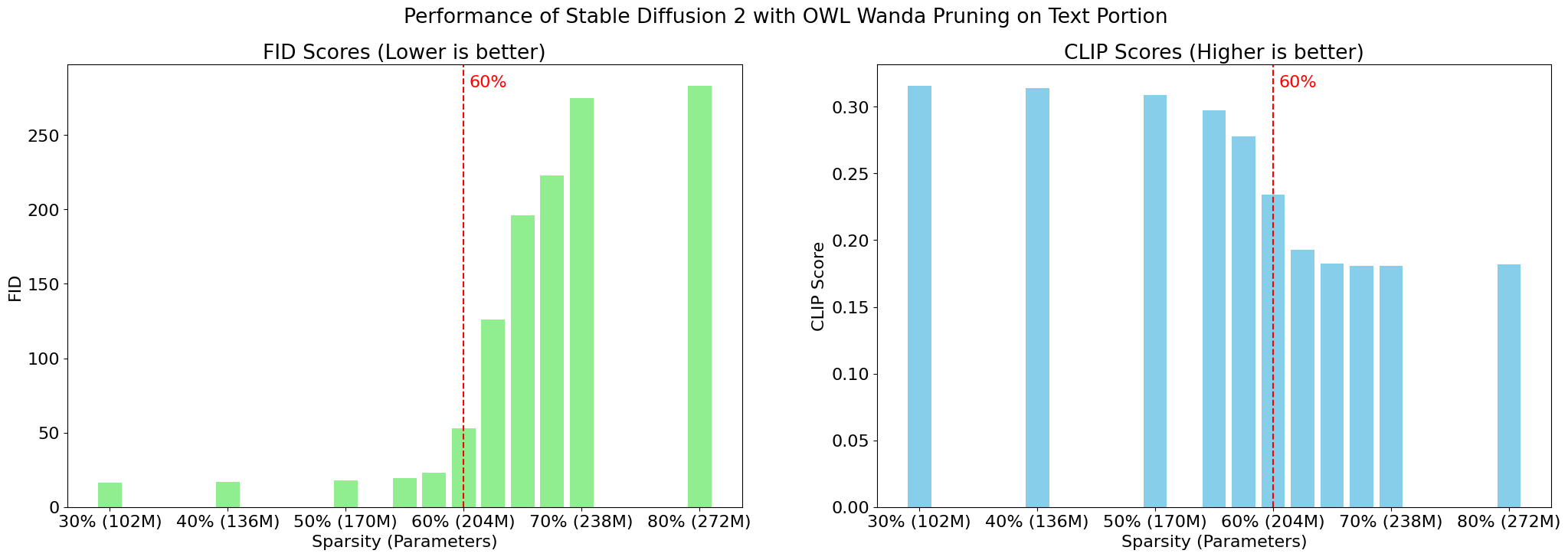}
    \caption{Pruning only Text Encoder of Stable Diffusion 2 using Wanda Pruning with OWL - Sharp drop off in performance observed at 60\%}
    \label{fig:sd2_text_owl_wanda_only_plots}
\end{figure}

We experiment with two pruning methods on the CLIP text encoder and extend these by applying OWL to test whether outlier weighting improves performance in text-to-image models. Figures~\ref{fig:sd2_text_magnitude_only_plots}, \ref{fig:sd2_text_wanda_only_plots}, \ref{fig:sd2_text_owl_magnitude_only_plots}, \ref{fig:sd2_text_owl_wanda_only_plots} show the FID and CLIP scores for each method. At lower sparsity levels, all techniques result in minimal performance degradation. However, as sparsity increases, each method experiences a sharp performance decline at specific thresholds. Qualitative evaluations of the generated images corroborate these quantitative results.

% When we pruned only the text encoder, without touching the image generator, we observed minimal loss of performance at low sparsities. Surprisingly, we found that there is a sharp drop-off in the quality of generated images at a specific sparsity threshold instead of the linear deterioration in performance that we had expected. This sharp drop-off occurs for all of the different pruning approaches we implemented. We confirmed this qualitative observation with both evaluation metrics used.

\paragraph{Magnitude Pruning}
% As the most simple and naive approach, magnitude pruning provides a valuable baseline. When pruning up to a sparsity of 60\%, the model performs very similarly to the base model and has a minimal drop in performance. However, at 62.5\%, the model performance drops suddenly. Looking at the images generated further reinforces this observation. The models seem to be losing the ability to understand the prompt at 62.5\%, as seen from the misframed dog and the lack of a field. After 62.5\% the model performs plummets and the images are now complete noise. 

As the simplest and most straightforward method, magnitude pruning serves as an important baseline. When pruning up to a sparsity of 60\%, the model performs similarly to the original, with only a minimal decline in performance. However, at 62.5\% sparsity, the model experiences a sudden drop in performance. This is further confirmed by visual inspection of the generated images, which show a clear degradation in quality. At 62.5\%, the model struggles to correctly interpret the prompt, as evidenced by a misframed dog and the absence of a field in the images. Beyond this threshold, the model's performance deteriorates drastically, with the generated images becoming complete noise.

\begin{figure}
    \centering
    \subfloat[0\%]{
        \includegraphics[width=0.225\textwidth]{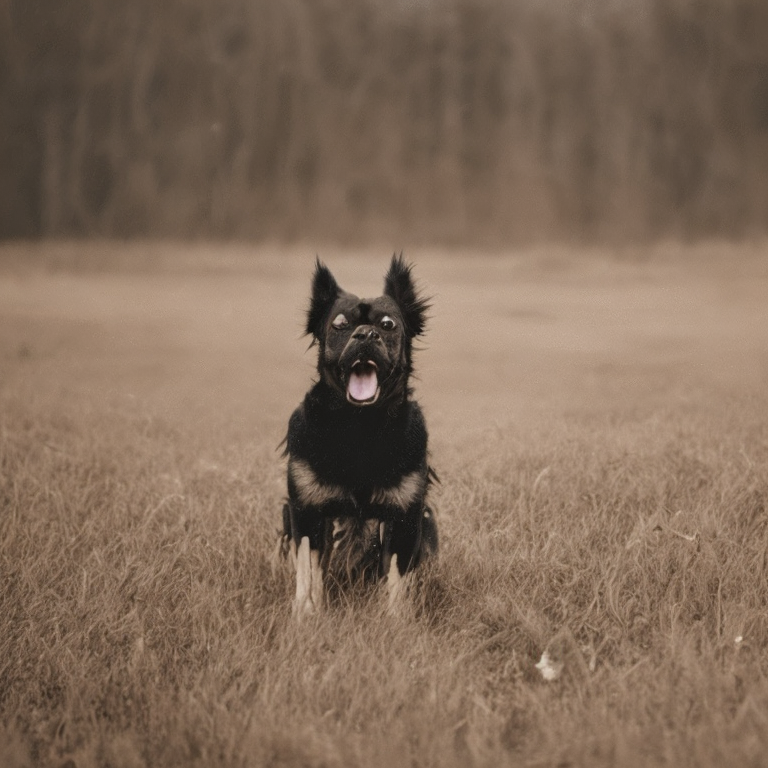}
    }
    \subfloat[50\%]{
        \includegraphics[width=0.225\textwidth]{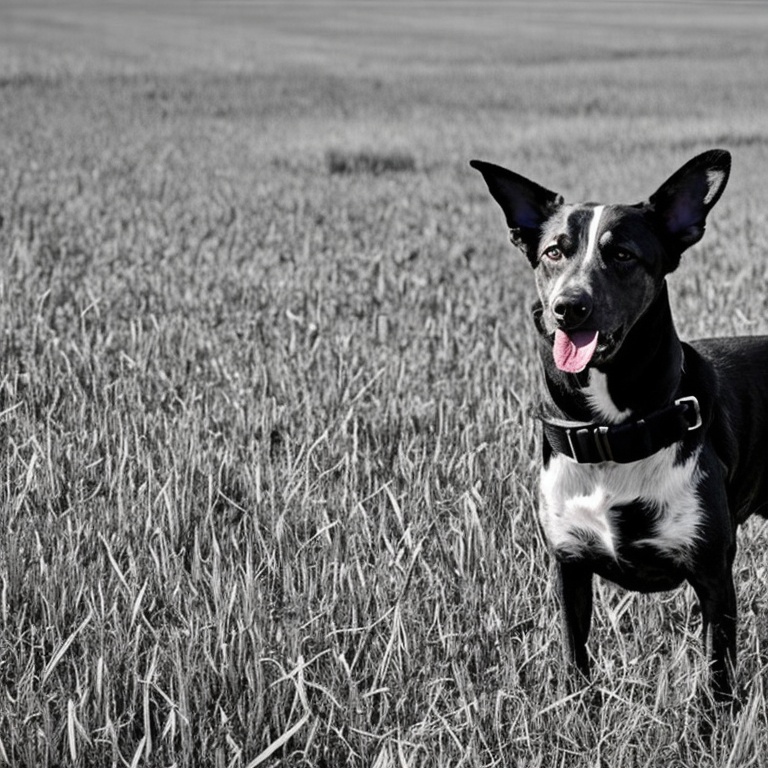}
    }
    
    \subfloat[62.5\%]{
        \includegraphics[width=0.225\textwidth]{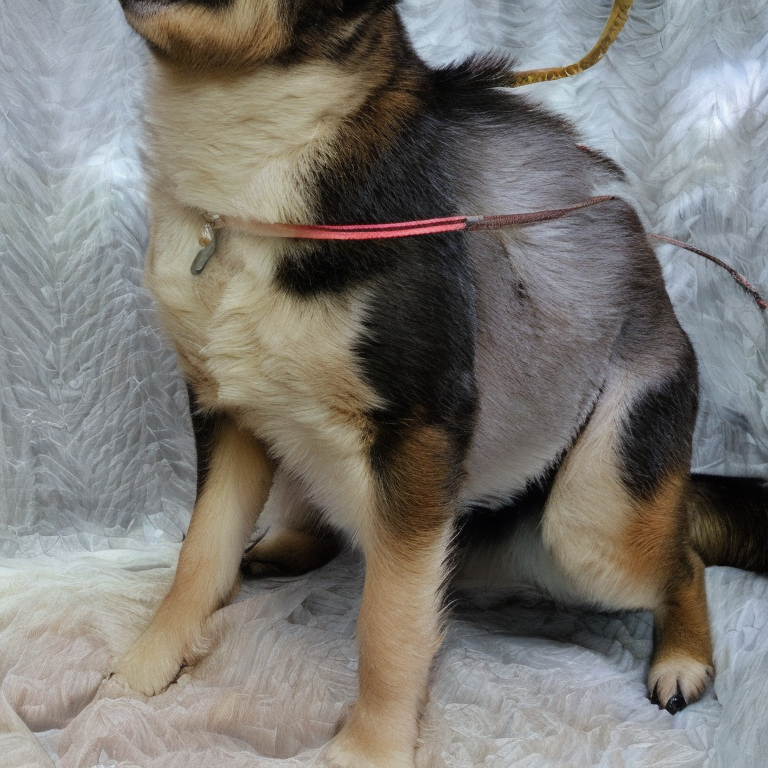}
    }
    \subfloat[65\%]{
        \includegraphics[width=0.225\textwidth]{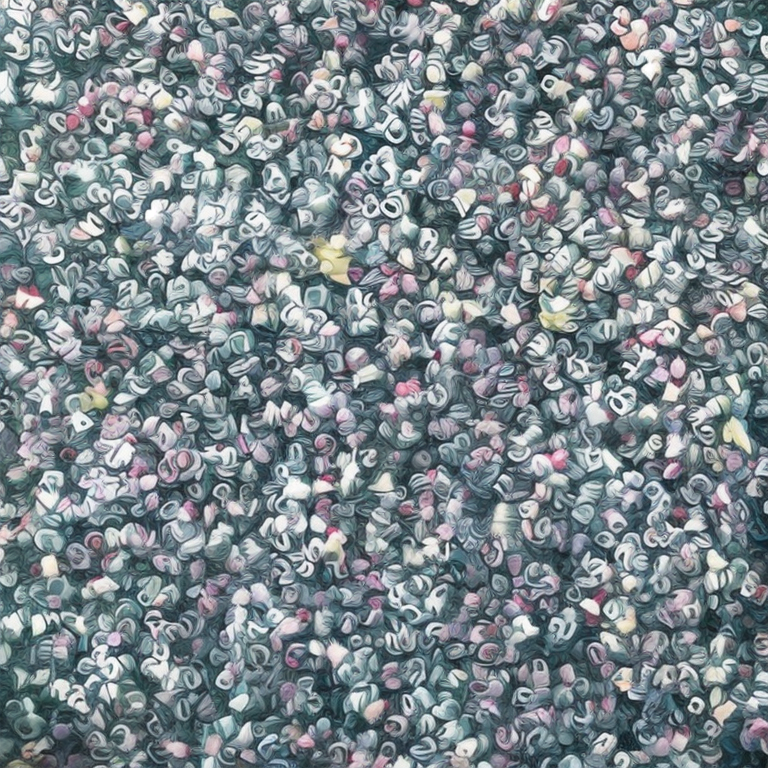}
    }
    \caption{Text Encoder Magnitude Pruning Examples - Text Encoder breaks down beyond 62.5\%}
    \label{fig:magnitude_only_pruning_examples}
\end{figure}

\paragraph{Wanda Pruning}
% Despite being the superior pruning technique for most language models, Wanda seems to be ill-suited to pruning the text component of this text-to-image model. When compared to the baseline of magnitude pruning, Wanda has very similar behavior but strictly worse performance. Wanda also has very little effect on performance up to a certain threshold and then causes a sharp drop in performance beyond this threshold. However, for Wanda, this threshold seems to be 60\%, meaning it is inferior to magnitude pruning.

% An interesting observation is that for the final experiment at 80\% sparsity, the FID is lower than seen at 70\%. This suggests that Wanda might perform better at higher sparsities, which is unexpected and requires further exploration. Another point of interest is that when looking at the images in Figure \ref{fig:wanda_only_pruning_examples}, the threshold seems slightly different from what is observed in the FID and CLIP Score graphs. The deterioration threshold seems to be 62.5\% rather than the 60\% seen in the metrics. This is likely the behavior for this specific prompt, while the metrics are calculated across 10,000 distinct prompts and capture the behavior on average.

Despite being one of the most effective pruning techniques for many language models, Wanda appears to be poorly suited for pruning the text component in this text-to-image model. When compared to the baseline magnitude pruning, Wanda exhibits similar behavior but consistently underperforms. Like magnitude pruning, Wanda shows minimal impact on performance up to a certain threshold, after which there is a sharp decline. However, for Wanda, this threshold occurs at 60\% sparsity, making it less effective than magnitude pruning.

\begin{figure}
    \centering
    \subfloat[0\%]{
        \includegraphics[width=0.225\textwidth]{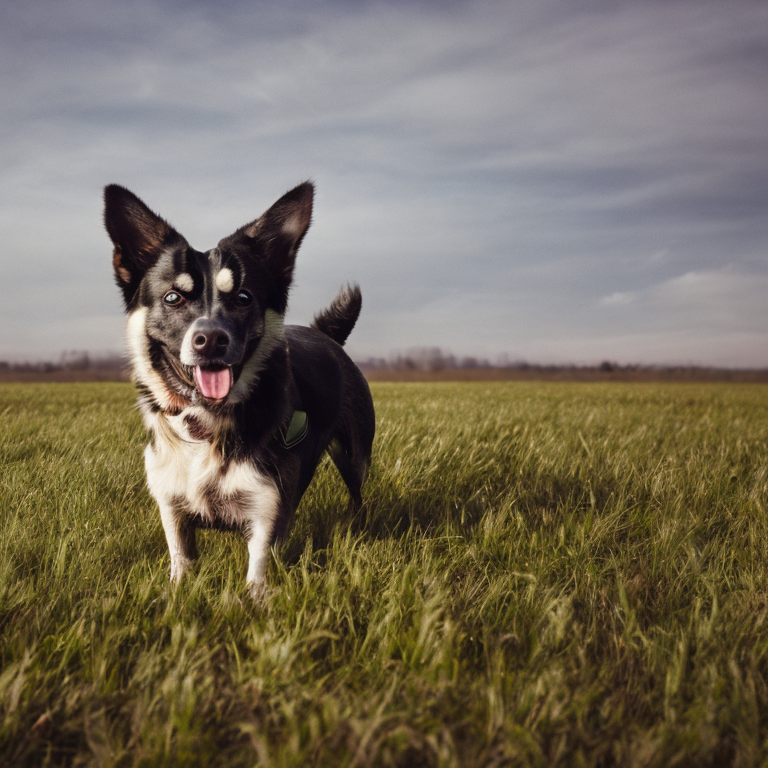}
    }
    \subfloat[50\%]{
        \includegraphics[width=0.225\textwidth]{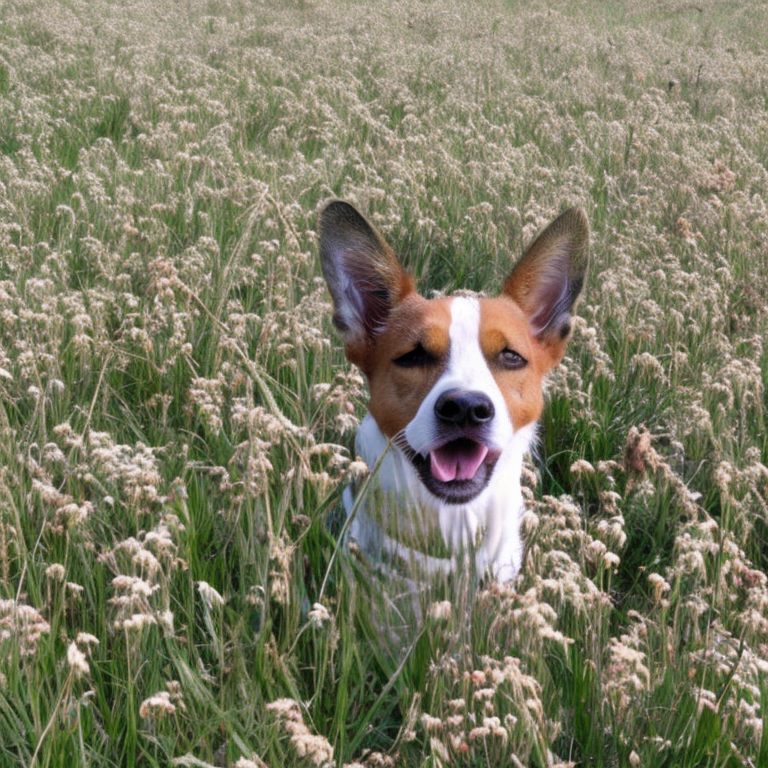}
    }
    
    \subfloat[62.5\%]{
        \includegraphics[width=0.225\textwidth]{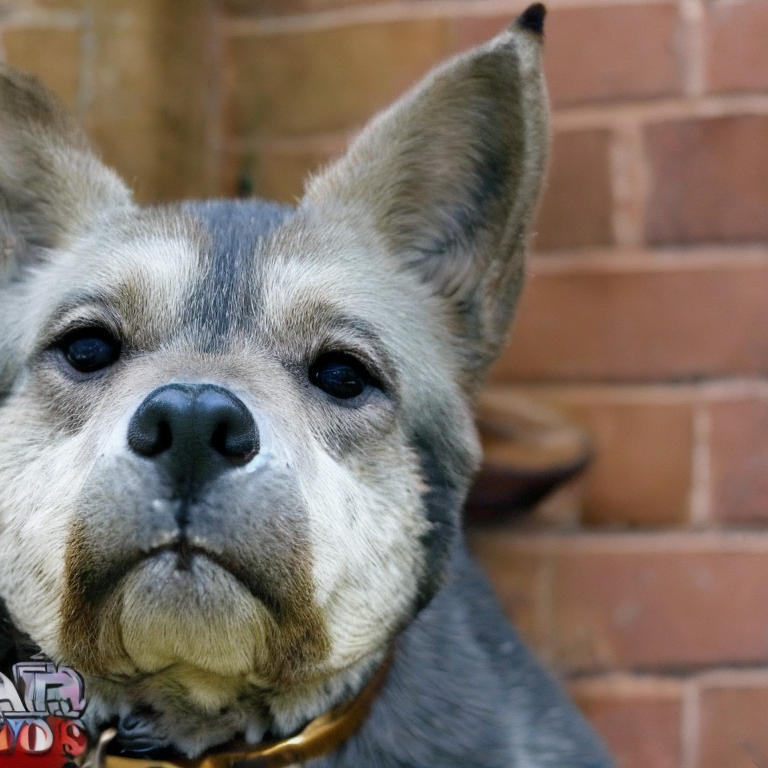}
    }
    \subfloat[65\%]{
        \includegraphics[width=0.225\textwidth]{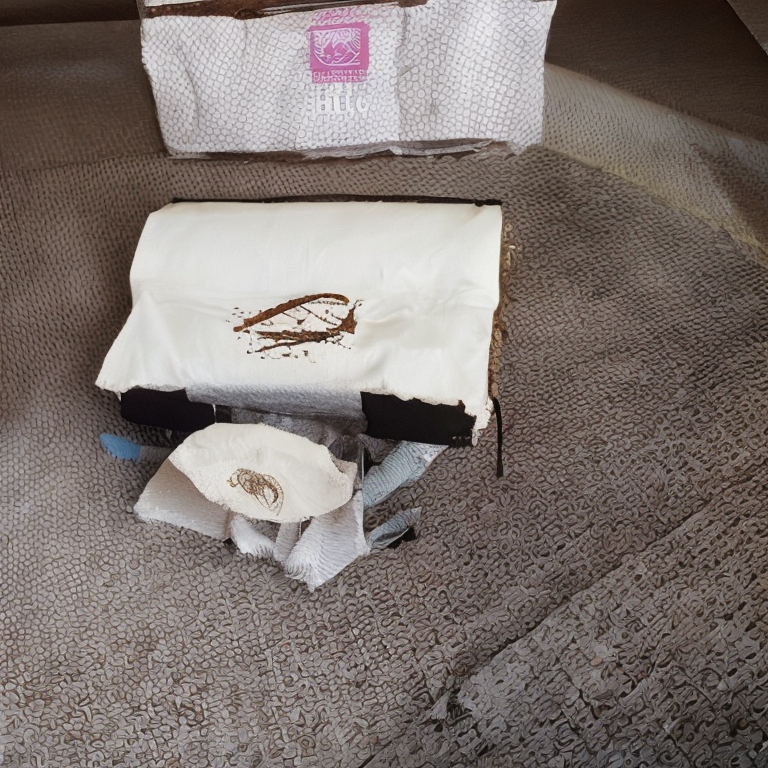}
    }
    \caption{Text Encoder Wanda Pruning Examples - Text Encoder breaks down beyond 62.5\%}
    \label{fig:wanda_only_pruning_examples}
\end{figure}

An interesting observation arises at 80\% sparsity, where the FID score is unexpectedly lower than at 70\%, suggesting that Wanda might perform better at higher sparsities. This anomaly warrants further investigation. Additionally, when visually inspecting the images in Figure \ref{fig:wanda_only_pruning_examples}, the deterioration threshold appears to be around 62.5\%, slightly different from the 60\% threshold indicated by the FID and CLIP Score metrics. This discrepancy is likely due to the specific behavior of the model on this particular prompt, whereas the metrics reflect the average performance across 10,000 distinct prompts.

\paragraph{Magnitude Pruning with OWL}
Applying outlier weighting to magnitude pruning shows minimal impact overall. While some sparsities exhibit slight improvements, the differences are generally insignificant. This finding aligns with the observations made by the authors of the OWL paper, who noted that outlier weighting tends to offer meaningful improvements primarily at higher sparsity levels.

\begin{figure}
    \centering
    \subfloat[0\%]{
        \includegraphics[width=0.225\textwidth]{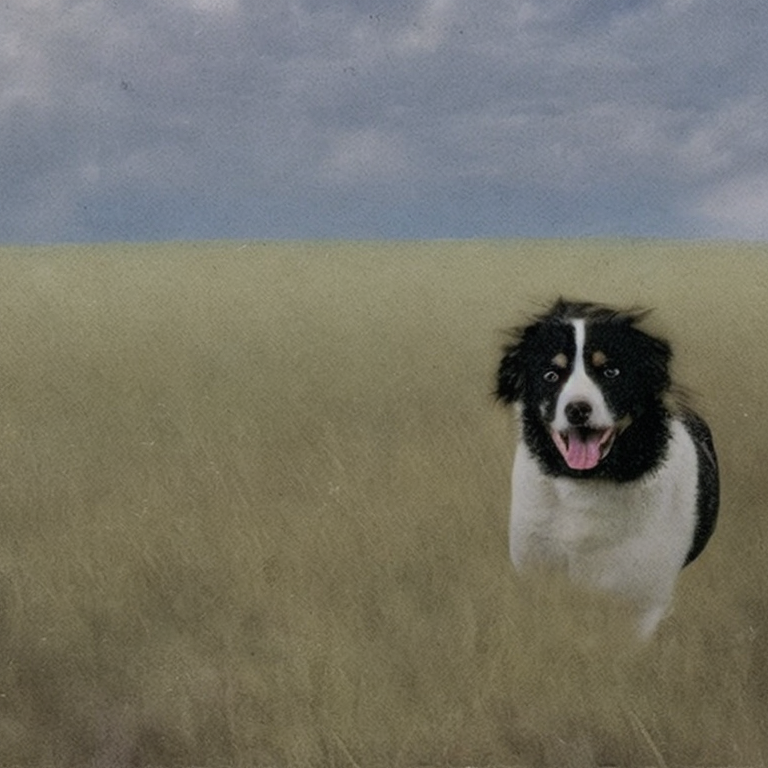}
    }
    \subfloat[50\%]{
        \includegraphics[width=0.225\textwidth]{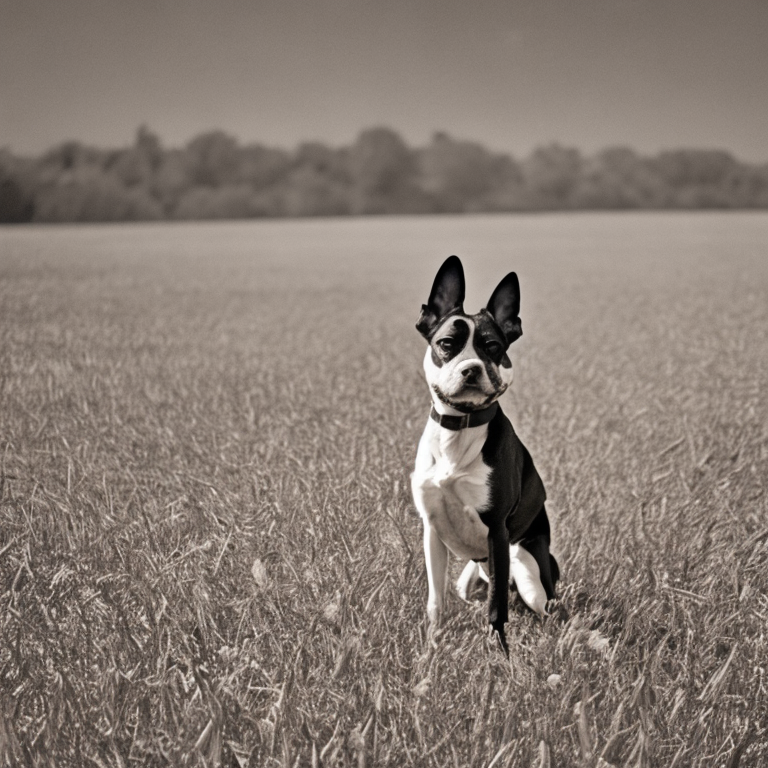}
    }
    
    \subfloat[62.5\%]{
        \includegraphics[width=0.225\textwidth]{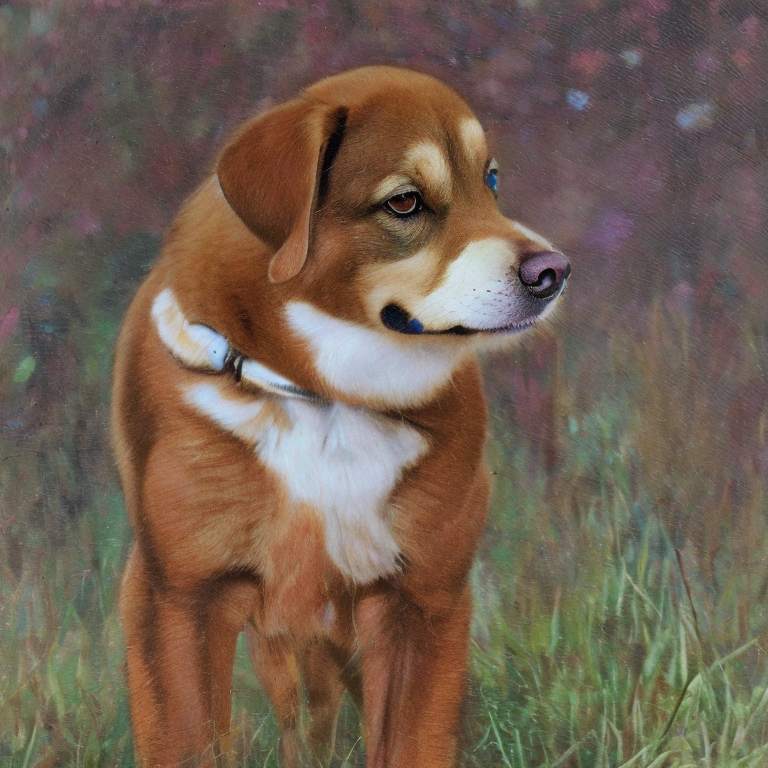}
    }
    \subfloat[65\%]{
        \includegraphics[width=0.225\textwidth]{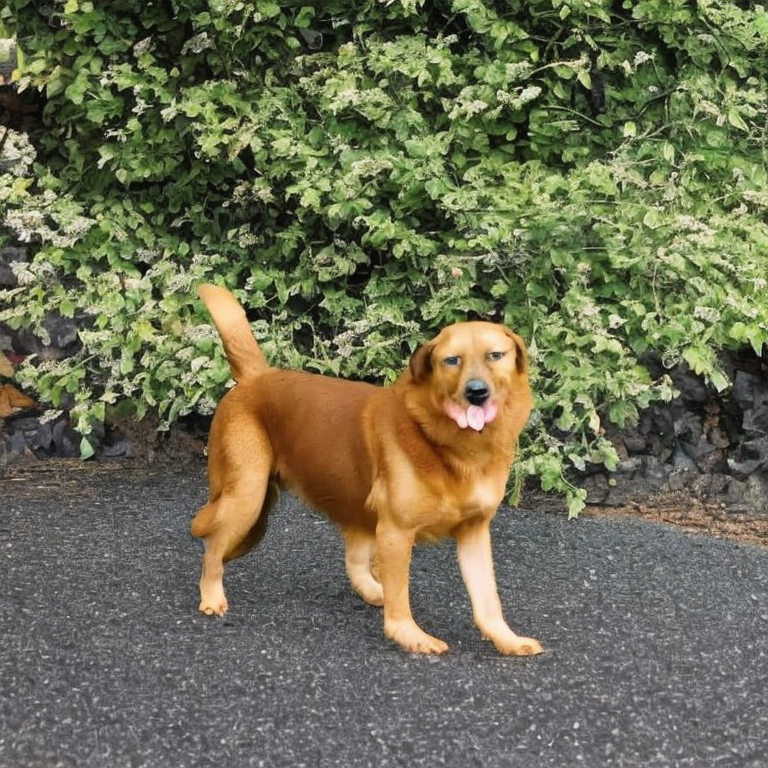}
    }
    \caption{Text Encoder Magnitude Pruning with OWL Examples - Text Encoder breaks down beyond 62.5\%}
    \label{fig:owl_magnitude_only_pruning_examples}
\end{figure}

\paragraph{Wanda Pruning with OWL}
As with magnitude pruning, applying outlier-based weighting to Wanda pruning does not result in significant improvements.

\begin{figure}
    \centering
    \subfloat[0\%]{
        \includegraphics[width=0.225\textwidth]{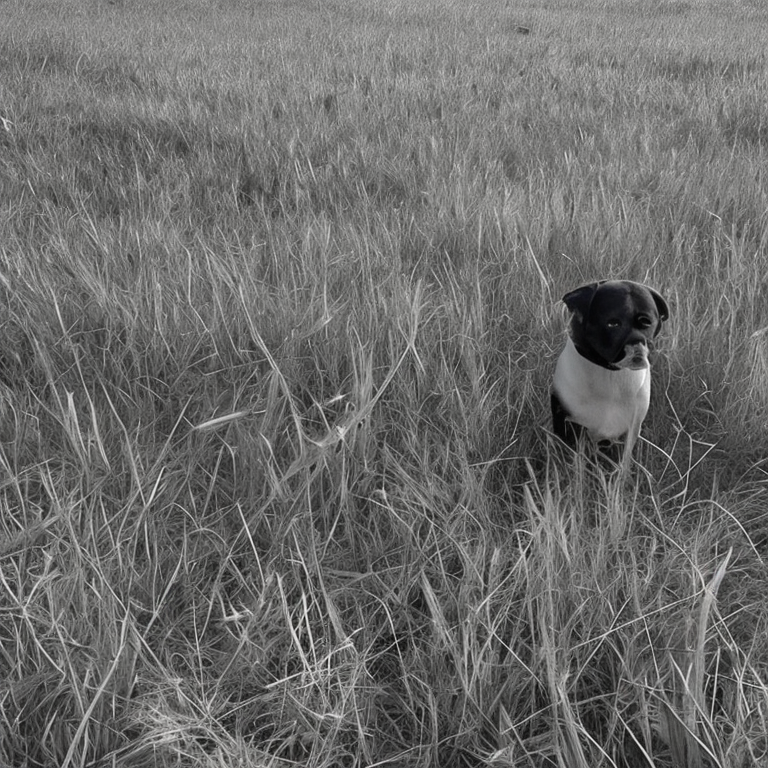}
    }
    \subfloat[50\%]{
        \includegraphics[width=0.225\textwidth]{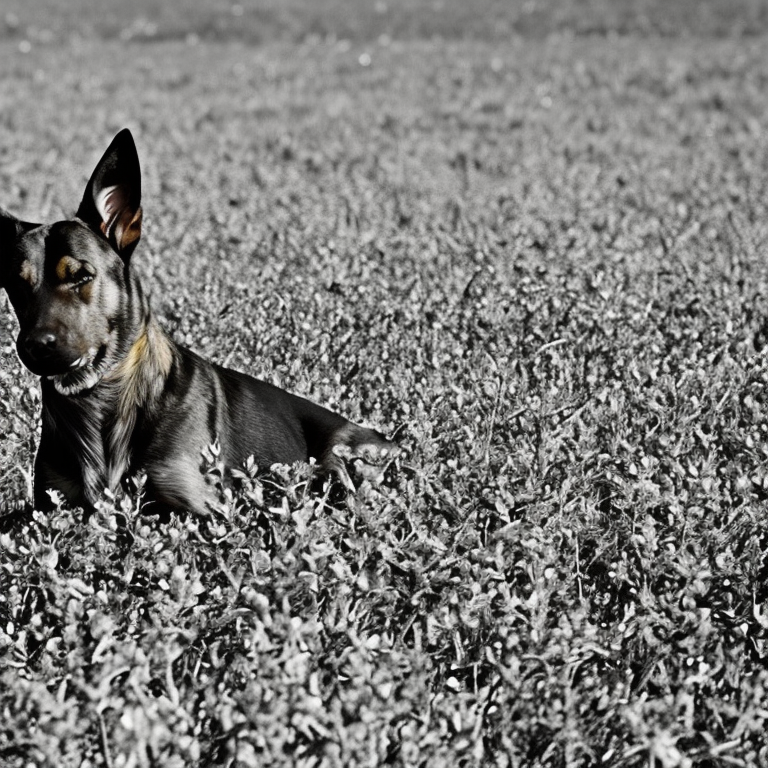}
    }
    
    \subfloat[62.5\%]{
        \includegraphics[width=0.225\textwidth]{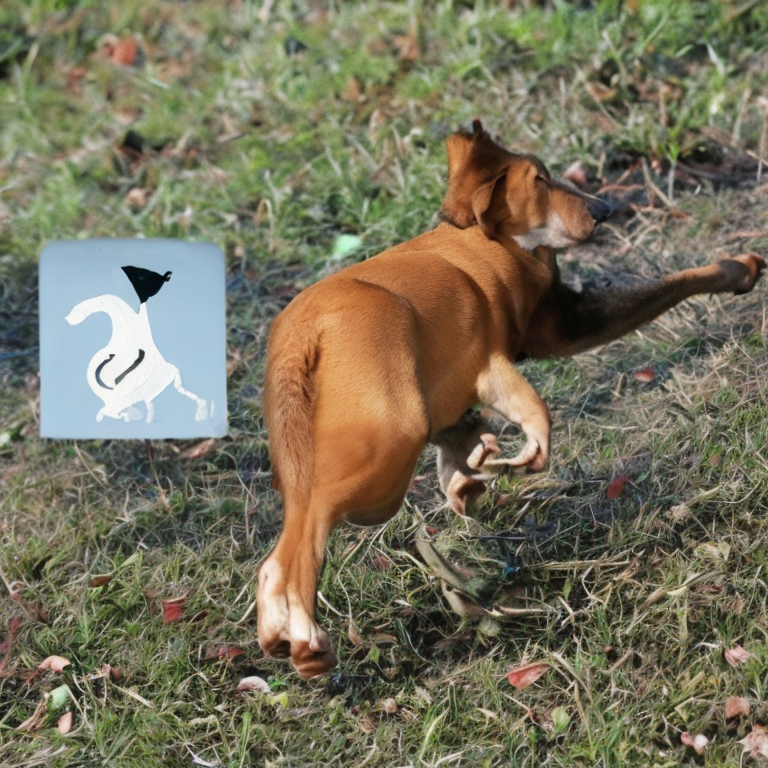}
    }
    \subfloat[65\%]{
        \includegraphics[width=0.225\textwidth]{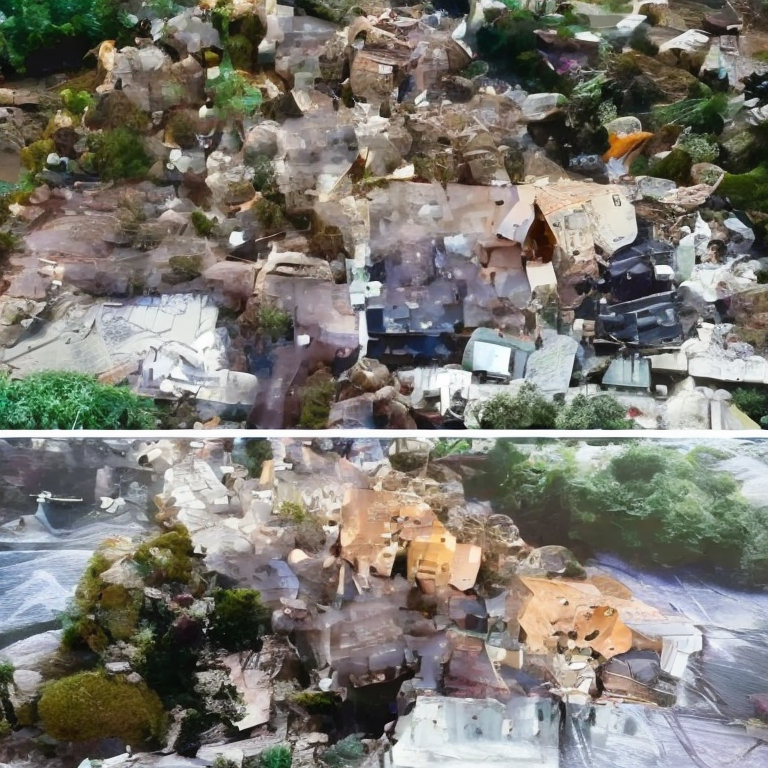}
    }
    \caption{Text Encoder Wanda Pruning with OWL Examples - Text Encoder breaks down beyond 60\%}
    \label{fig:owl_wanda_only_pruning_examples}
\end{figure}

\paragraph{Comparison of Pruning Techniques}
% The observation that Wanda pruning is outperformed by magnitude pruning is quite surprising and strongly indicates the need for more specific pruning techniques for text-to-image models. As seen in Figure \ref{fig:pruning_comparions_line_plot}, at any given sparsity, magnitude outperforms Wanda. For other NLP models, magnitude pruning typically performs the worst when compared to other post-training pruning techniques, so there is significant scope for developing bespoke algorithms for post-training pruning these models. The other observation is that Outlier Weighted Layerwise pruning on top of magnitude and Wanda pruning provides marginal improvement. Figure \ref{fig:pruning_comparions_line_plot} once again highlights the sudden drop off observed irrespective of the pruning algorithm. 

\begin{figure}
    \centering
    \includegraphics[width=\linewidth]{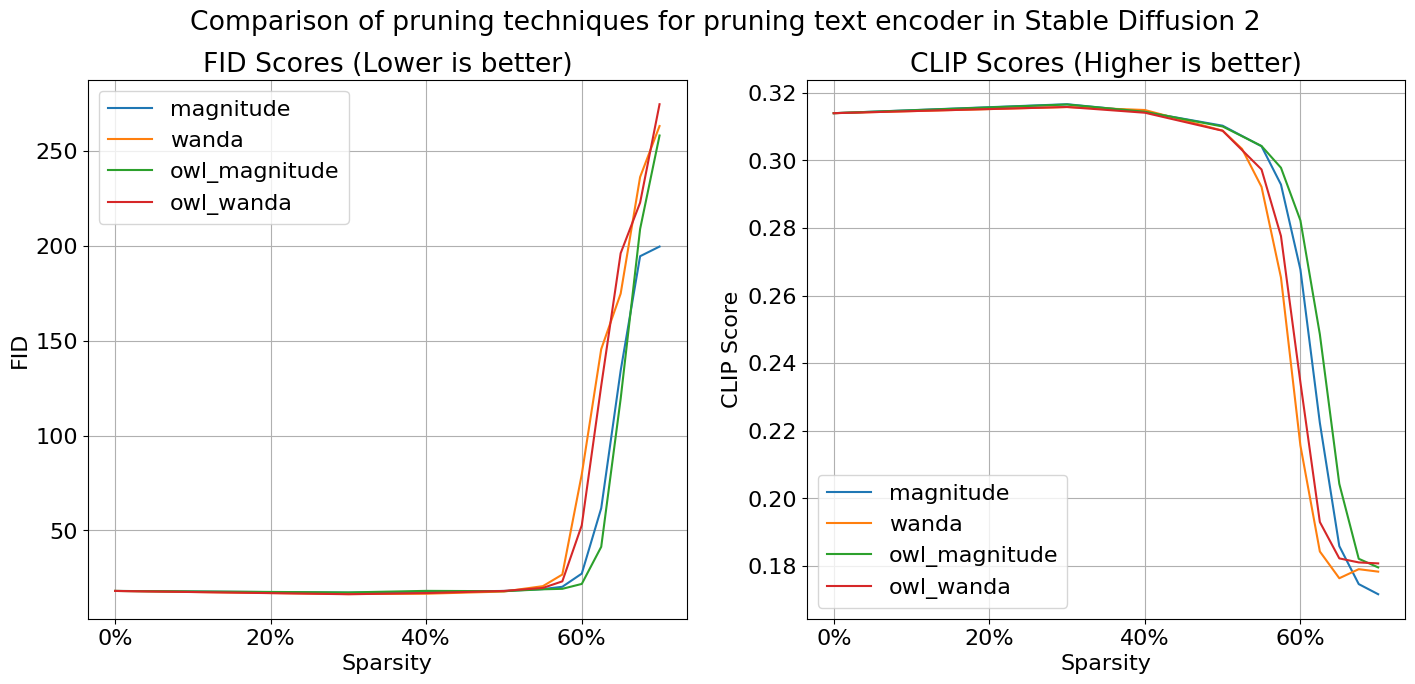}
    \caption{Comparing Pruning Techniques for pruning only text encoder}
    \label{fig:pruning_comparions_line_plot}
\end{figure}

The finding that Wanda pruning is outperformed by magnitude pruning is quite surprising and underscores the need for pruning techniques tailored specifically to text-to-image models. As shown in Figure \ref{fig:pruning_comparions_line_plot}, magnitude pruning consistently outperforms Wanda at every sparsity level. This is particularly notable given that, in most NLP models, magnitude pruning is typically the weakest among post-training pruning methods. This suggests a significant opportunity to develop specialized pruning algorithms for text-to-image models.

Another key observation is that Outlier Weighted Layerwise (OWL) pruning, when applied on top of both magnitude and Wanda pruning, results in only marginal improvements. Figure \ref{fig:pruning_comparions_line_plot} also highlights the sharp performance drop-off, a pattern observed across all pruning algorithms tested.

\subsection{U-Net Pruning Only}
% For pruning the diffusion component of the model, we observed a more gradual decrease in performance. The pruned model continued to perform well at lower sparsities, but the performance dropped steadily as the sparsity increased above 40\%. Looking at the results qualitatively, 50\% was chosen as the point at which the model broke down, and this threshold was used in full model pruning. This gradual deterioration in performance is clear when viewing all pruned images from 30\% to 60\%, as is shown in Figure \ref{fig:diffusion_magnitude_only_gradual_shift} in the appendix.

For the diffusion component of the model, we observe a more gradual decline in performance as pruning increased. The model maintains strong performance at lower sparsity levels, but its performance steadily decreases as sparsity exceeded 40\%. Based on a qualitative analysis, 50\% sparsity is identified as the point where the model's performance significantly deteriorated, and this threshold is later used in full model pruning. This gradual decline is evident when examining the pruned images from 30\% to 60\%, as shown in Figure \ref{fig:diffusion_magnitude_only_gradual_shift} in the appendix.

\begin{figure}
    \centering
    \includegraphics[width=\linewidth]{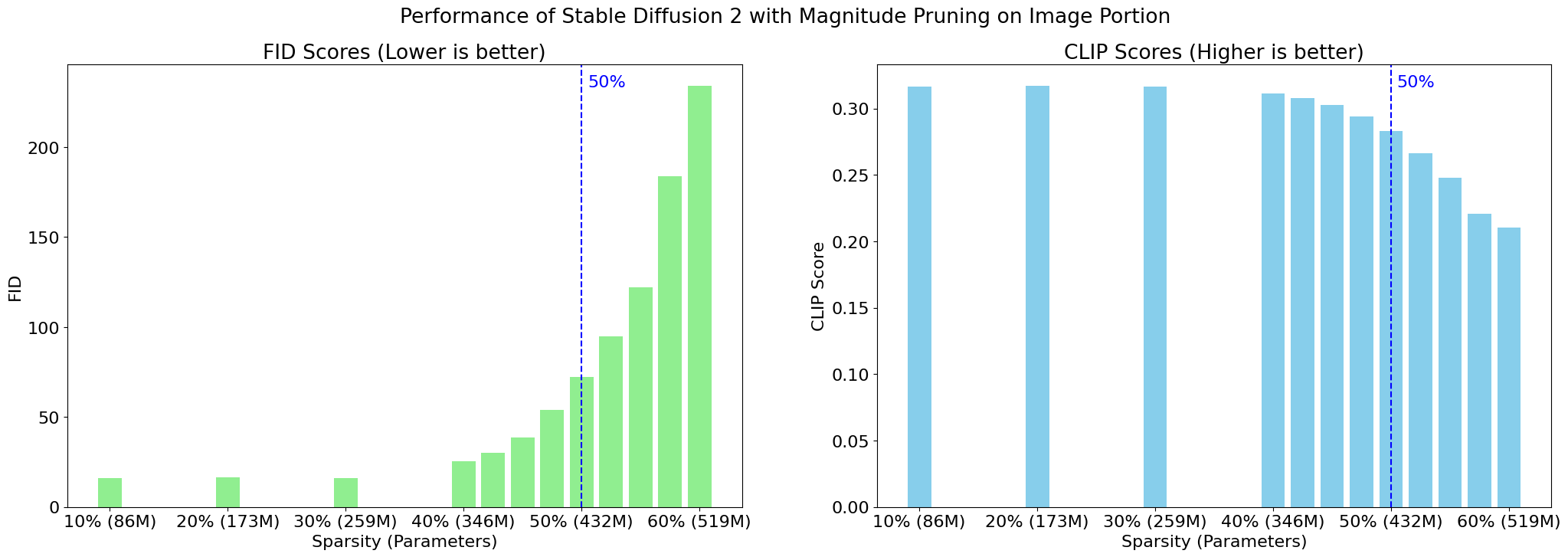}
    \caption{Pruning only Image Diffusion Generator of Stable Diffusion 2 using Magnitude Pruning - Gradual decline in performance}
    \label{fig:sd2_image_magnitude_only_plots}
\end{figure}

\begin{figure}
    \centering
    \subfloat[0\%]{
        \includegraphics[width=0.225\textwidth]{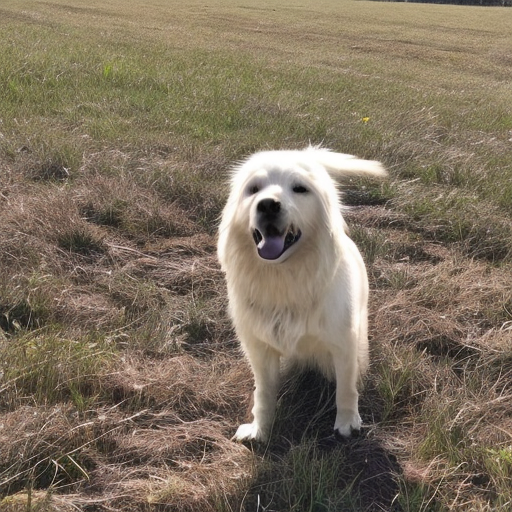}
    }
    \subfloat[40\%]{
        \includegraphics[width=0.225\textwidth]{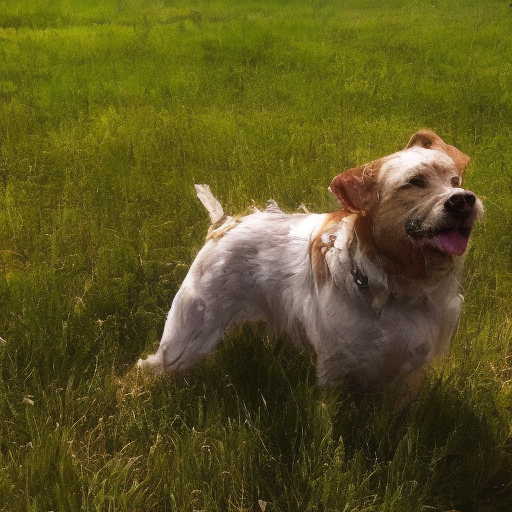}
    }
    
    \subfloat[50\%]{
        \includegraphics[width=0.225\textwidth]{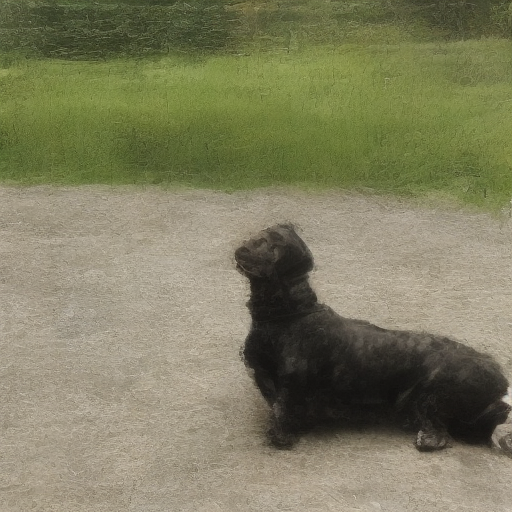}
    }
    \subfloat[60\%]{
        \includegraphics[width=0.225\textwidth]{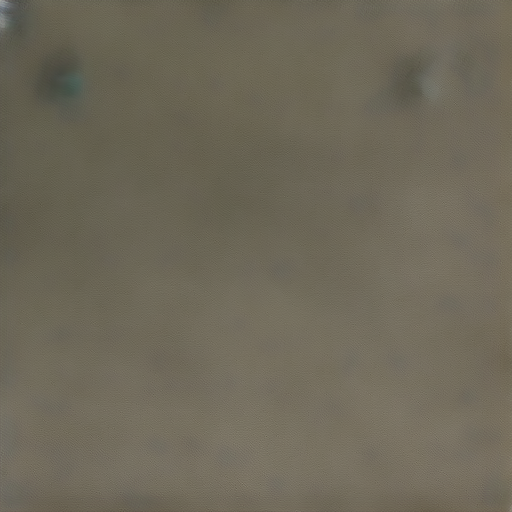}
    }
    \caption{Diffusion Generator Magnitude Pruning Examples - Performance greatly suffers beyond 40\%}
    \label{fig:diffusion_magnitude_only_pruning_examples}
\end{figure}

\subsection{Comparison between Text and Image Pruning}

% To answer the question of which between the text encoder and image diffusion generator yields better results when pruned, we look at a comparison of the best-pruned models from both approaches in Table \ref{tab:text_image_comparison}.

To address the question of which component—the text encoder or the image diffusion generator—yields better results when pruned, we compare the best-performing pruned models for each approach. This comparison is presented in Table \ref{tab:text_image_comparison}.

\begin{table}
    \centering
    \scalebox{0.9}{
    \begin{tabular}{cccccc}
        \hline
        Total Sparsity & Text Sparsity & Image Sparsity & FID & CLIP Score\\
        \hline
        16.9\% & 60\% & 0\% & \cellcolor{green!25}21.7 & 0.282 \\
        28.7\% & 0\% & 40\% & 25.38 & \cellcolor{green!25}0.311 \\
        \hline
    \end{tabular}
    }
    \caption{Comparison between best model obtained pruning each component individually}
    \label{tab:text_image_comparison}
\end{table}

% Comparing these two models that qualitatively produce similar results, we observe some interesting results. Firstly it seems that pruning the image diffusion generator yields better sparsity since it is larger and results in a greater full model sparsity. The other observation is that pruning the text encoder results in a worse CLIP Score and pruning the image diffusion generator results in a worse FID. This leads to some interesting questions about what these metrics are measuring. In particular, it is possible that the CLIP Score is worse for text encoder pruning because the text encoder model is also a CLIP model and this introduces some bias. 

Comparing the two models, which qualitatively produce similar results, reveals some intriguing findings. First, pruning the image diffusion generator achieves greater overall sparsity due to its larger size, contributing more significantly to the full model’s sparsity. Additionally, pruning the text encoder results in a worse CLIP score, while pruning the image diffusion generator leads to a worse FID score. This raises interesting questions about what these metrics are actually measuring. In particular, the lower CLIP score for text encoder pruning may be due to a potential bias, as the text encoder itself is a CLIP model, which could influence how its performance is evaluated.

\subsection{Full Model Pruning}

\paragraph{First Approach}

Given the computational demands of evaluating models, conducting an exhaustive grid search to identify the optimal pruning configuration is impractical. In the first approach we try to establish guiding principles for how to best distribute sparsity given a certain target sparsity for the full model. The experiments shown in Table \ref{tab:full_model_sparsities_table} were conducted and the results are shown in Tables \ref{tab:full_model_fid_scores}, \ref{tab:full_model_clip_scores}, \ref{tab:full_model_sparsities_results}. As outlined in the methodology section, certain configurations are deemed invalid due to the significant size disparity between the text and image components. 

All experiments used magnitude pruning for both components of the model. Similar experiments were conducted using magnitude pruning with OWL for the CLIP text encoder and produced very similar results as seen in  Tables \ref{tab:owl_full_model_fid_scores}, \ref{tab:owl_full_model_clip_scores}, \ref{tab:owl_full_model_sparsities_results} in the appendix. 

% There were several approaches that we considered for how to best find the optimal full model pruning configuration. Due to the large amount of time required to generate the 10,000 images necessary for computing evaluation metrics, an exhaustive grid search was infeasible. So instead we tried to answer the question of what is the best way of distributing the sparsity given a specific full model sparsity. The experiments show in Table \ref{tab:full_model_sparsities_table} were conducted and the results are shown in Tables \ref{tab:full_model_fid_scores}, \ref{tab:full_model_clip_scores}, \ref{tab:full_model_sparsities_results}. As discussed in the methodology section, certain configurations are not valid because the text portion is significantly smaller than the image portion. 

Tables \ref{tab:full_model_fid_scores}, \ref{tab:full_model_clip_scores}, \ref{tab:full_model_sparsities_results} demonstrate that the model performs better when the majority of the sparsity is allocated to the image component. Notably, the configuration in which 75\% of the model's sparsity is concentrated in the image diffusion generator consistently outperforms other splits. This finding aligns with the earlier observation that pruning the image component yields greater sparsity with less decline in performance compared to pruning the text component.

The models where performance deteriorates significantly are highlighted in Table \ref{tab:full_model_sparsities_results} and correspond to sparsities that exceed the drop-off thresholds identified in the individual component pruning experiments. This reinforces the consistency of these pruning thresholds and informs our second approach for determining the optimal full model pruning configuration.

% The models where performance completely breaks down are highlighted in Table \ref{tab:full_model_sparsities_results} and occur when sparsities violate the drop-off thresholds found in the individual component pruning experiments. This reinforces consistency of these pruning thresholds and lead into the second approach for findinf the optimal full model pruning configuration.

% For this section, all experiments pruned the text portion of the model using both magnitude pruning and magnitude pruning with OWL, the two best techniques. The results presented in Tables \ref{tab:full_model_fid_scores}, \ref{tab:full_model_clip_scores}, \ref{tab:full_model_sparsities_results} are from the experiments using magnitude pruning. The magnitude pruning with OWL showed very similar results which can be seen in Tables \ref{tab:owl_full_model_fid_scores}, \ref{tab:owl_full_model_clip_scores}, \ref{tab:owl_full_model_sparsities_results} in the appendix. 

% Looking at these results there are a few observations to be made. In general, it seems that models perform better when their sparsity is mostly in the image portion of the model as highlighted in Table \ref{tab:full_model_fid_scores} and Table \ref{tab:full_model_clip_scores}. When comparing these splits of sparsity, the split where 75\% of the model's sparsity is in the image portion performs consistently better. This observation agrees with the previously observed result that pruning the image portion over the text portion yields more sparsity for less drop in performance.

\begin{table}
    \centering
    \begin{tabular}{c|ccc}
        \hline
        Total Sparsity & 75:25 & 50:50 & \cellcolor{green!25}25:75 \\
        \hline
        20\% & 20.92 & 20.89 & \cellcolor{green!25}17.85 \\
        30\% & 244.92 & 20.8 & \cellcolor{green!25}18.4 \\
        40\% & N/A & 219.97 & \cellcolor{green!25}20.71 \\
        50\% & N/A & 202.84 & \cellcolor{green!25}66.2 \\
        60\% & N/A & N/A & \cellcolor{green!25}370.2 \\
        \hline
    \end{tabular}
    \caption{FID at different allocations of sparsity between text encoder and image diffusion generator}
    \label{tab:full_model_fid_scores}
\end{table}

\begin{table}
    \centering
    \begin{tabular}{c|ccc}
        \hline
        Total Sparsity & 75:25 & 50:50 & \cellcolor{green!25}25:75 \\
        \hline
        20\% & 0.308 & 0.314 & \cellcolor{green!25}0.318 \\
        30\% & 0.169 & 0.307 & \cellcolor{green!25}0.316 \\
        40\% & N/A & 0.176 & \cellcolor{green!25}0.307 \\
        50\% & N/A & 0.181 & \cellcolor{green!25}0.27 \\
        60\% & N/A & N/A & \cellcolor{green!25}0.202 \\
        \hline
    \end{tabular}
    \caption{CLIP Score at different allocations of sparsity between text encoder and image diffusion generator}
    \label{tab:full_model_clip_scores}
\end{table}

% Another interesting observation from these results highlighted in Table \ref{tab:full_model_sparsities_results} is that some models perform significantly worse than the base model. These pruning configurations all have sparsities that violate the thresholds found in the individual pruning section. 

\begin{table}
    \centering
    \scalebox{0.75}{
    \tabcolsep=0.11cm
    \begin{tabular}{cccccc}
        \hline
        Total Sparsity & Text:Image Ratio & Text Sparsity & Image Sparsity & FID & CLIP Score\\
        \hline
        20\% & 75:25 & 53\% & 7\% & 20.92 & 0.308\\
        20\% & 50:50 & 35\% & 14\% & 20.89 & 0.314\\
        20\% & 25:75 & 18\% & 21\% & 17.85 & 0.318\\
        \hline
        30\% & 75:25 & \cellcolor{red!25}80\% & 10\% & \cellcolor{red!25}244.92 & \cellcolor{red!25}0.169\\
        30\% & 50:50 & 53\% & 21\% & 20.8 & 0.307\\
        30\% & 25:75 & 27\% & 31\% & 18.4 & 0.316\\
        \hline
        40\% & 50:50 & \cellcolor{red!25}71\% & 28\% & \cellcolor{red!25}219.97 & \cellcolor{red!25}0.176\\
        40\% & 25:75 & 35\% & 42\% & 20.71 & 0.307\\
        \hline
        50\% & 50:50 & \cellcolor{red!25}89\% & 35\% & \cellcolor{red!25}202.84 & \cellcolor{red!25}0.181\\
        50\% & 25:75 & 44\% & \cellcolor{red!25}52\% & \cellcolor{red!25}66.2 & \cellcolor{red!25}0.27\\
        \hline
        60\% & 25:75 & 53\% & \cellcolor{red!25}63\% & \cellcolor{red!25}370.2 & \cellcolor{red!25}0.202\\
        \hline
        
    \end{tabular}
    }
    \caption{Model Performance greatly suffers when component sparsities violate the identified drop-off thresholds}
    \label{tab:full_model_sparsities_results}
\end{table}

% These results reinforce the consistency of the pruning thresholds found during the individual component pruning section and lead to the second approach to full model pruning. If the models showed very little drop in performance below the threshold, the best configuration would be to prune both portions of the model up to the threshold. 

\paragraph{Second Approach}

% We tested the model when pruned to both thresholds found in the individual pruning experiments. It seemed likely that these thresholds might not behave the same when both portions were pruned together, so we also pruned several models to a certain amount less than the threshold, to find the optimal sparsities. Starting at the two thresholds, both the text portion sparsity and image portion sparsity are reduced in steps of 2.5\% and each of these models is also evaluated. These results are seen in Table \ref{tab:final_pruning_configs_results}. The unpruned model is provided as a baseline and highlighted in gray. All of these configurations were tested with both magnitude and magnitude with OWL for the text portions. The results shown in Table \ref{tab:final_pruning_configs_results} are for magnitude pruning. The results using magnitude pruning with OWl are largely the same and can be seen in the appendix in Table \ref{tab:final_pruning_configs_results_owl_version}.

We examine the model pruned to both thresholds established in the individual pruning experiments. Given the possibility that these thresholds might behave differently when both components are pruned simultaneously, we also prune several models to values slightly below the thresholds to determine the optimal sparsity configurations. Starting from the two thresholds, we decrease both the text and image component sparsities in increments of 2.5\%, evaluating each of these configurations. The results are summarized in Table \ref{tab:final_pruning_configs_results}, where the unpruned model is provided as a baseline and highlighted in gray. The models were pruned using magnitude pruning for both components of the model. These configurations were also tested using magnitude pruning with OWL for CLIP text encoder and produced very similar results as shown in Table \ref{tab:final_pruning_configs_results_owl_version} in the appendix

% All configurations were tested using both magnitude pruning and magnitude pruning with Outlier Weighted Layerwise (OWL) for the text components. The results displayed in Table \ref{tab:final_pruning_configs_results} correspond to magnitude pruning. The outcomes for magnitude pruning with OWL are largely consistent and can be found in the appendix in Table \ref{tab:final_pruning_configs_results_owl_version}.

\begin{figure}
    \centering
    \includegraphics[width=0.9\linewidth]{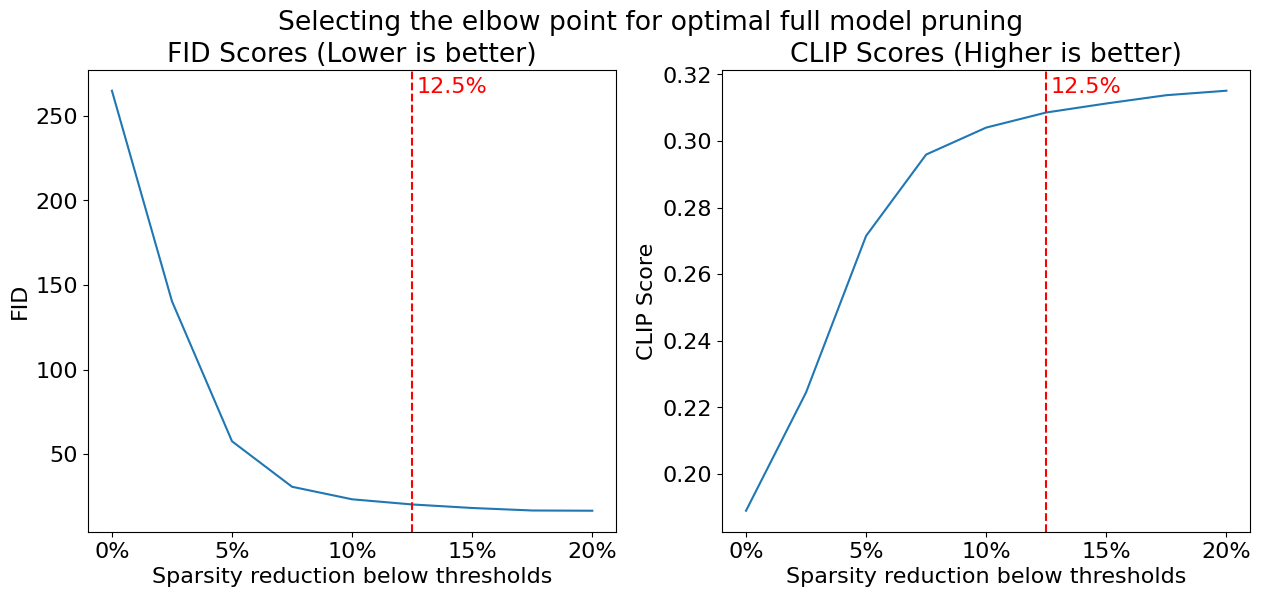}
    \caption{Model Improvement as sparsity is reduced from identified drop-off thresholds}
    \label{fig:model_reccomendation_knee}
\end{figure}

\begin{table}
    \centering
    \scalebox{0.9}{
    \begin{tabular}{ccccc}
        \hline
        Total Sparsity & Text Sparsity & Image Sparsity & FID & CLIP Score\\        
        \hline
        \cellcolor{gray!40}0\% & \cellcolor{gray!40}0\% & \cellcolor{gray!40}0\% & \cellcolor{gray!40}18.07 & \cellcolor{gray!40}0.314\\
        53.5\% & 62.5\% & 50.0\% & 264.87 & 0.189\\
        51.0\% & 60.0\% & 47.5\% & 140.39 & 0.224\\
        48.5\% & 57.5\% & 45.0\% & 57.59 & 0.272\\
        46.0\% & 55.0\% & 42.5\% & 30.69 & 0.296\\
        43.5\% & 52.5\% & 40.0\% & 23.28 & 0.304\\
        41.0\% & 50.0\% & 37.5\% & 20.24 & 0.309\\
        \cellcolor{green!25}38.5\% & \cellcolor{green!25}47.5\% & \cellcolor{green!25}35.0\% & \cellcolor{green!25}18.15 & \cellcolor{green!25}0.311\\
        36.0\% & 45.0\% & 32.5\% & 16.66 & 0.314\\
        33.5\% & 42.5\% & 30.0\% & 16.53 & 0.315\\
        \hline
    \end{tabular}
    }
    \caption{Pruning Stable Diffusion 2 to found thresholds}
    \label{tab:final_pruning_configs_results}
\end{table}

% Based on the results seen in Table \ref{tab:final_pruning_configs_results}, it seems that pruning to both thresholds together seems to result in a model that is much worse than expected. These thresholds individually caused minimal loss of performance, but it appears that pruning both parts together causes the model to perform significantly worse. As both sparsities are steadily lowered, we found an improvement and were able to use these results along with qualitative evaluation of the generated images to make a recommendation for the optimal pruning configuration for the model.

% Stable Diffusion 2 can be pruned to 38.5\% sparsity by pruning the text encoder to 47.5\% and the diffusion generator to 35\%. Pruning to this configuration causes minimal loss in the quality of generated images and provides a significant reduction in model size. A comparison of the generated image quality can be seen in Figure \ref{fig:base_vs_optimal_config}.

The results presented in Table \ref{tab:final_pruning_configs_results} indicate that pruning both thresholds simultaneously leads to a model performance that is significantly worse than anticipated. While each threshold individually resulted in minimal performance loss, the combined pruning of both components appears to degrade the model’s performance substantially. However, as we gradually reduce both sparsities, we observe improvements, which, along with qualitative evaluations of the generated images, enable us to recommend an optimal pruning configuration for the model.

\begin{figure}
    % Row 1
    \caption*{Prompt: A dog in a field}
    \begin{subfigure}{0.45\linewidth}
        \centering
        \includegraphics[width=\linewidth]{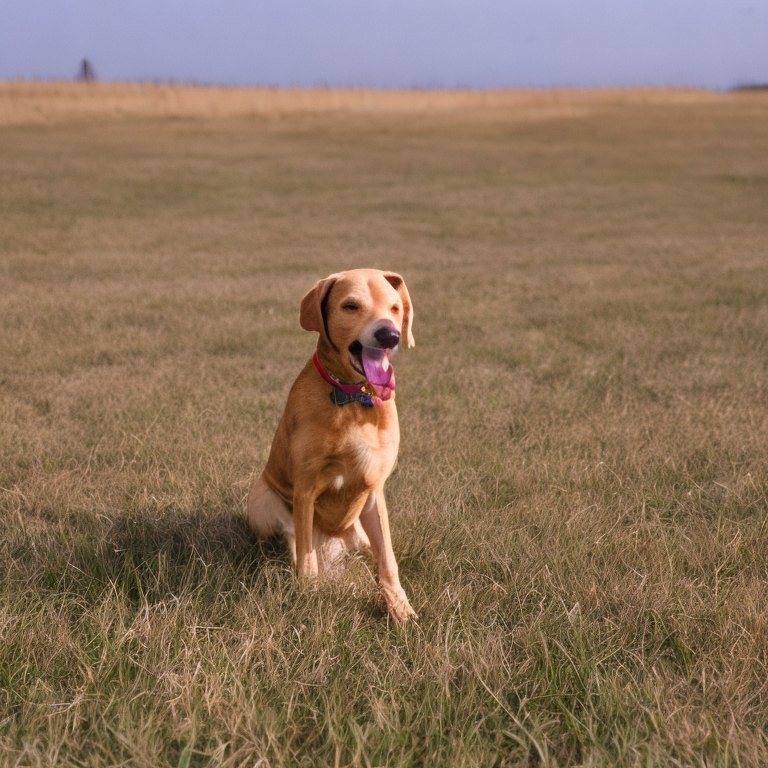}
    \end{subfigure}
    \hfill
    \begin{subfigure}{0.45\linewidth}
        \centering
        \includegraphics[width=\linewidth]{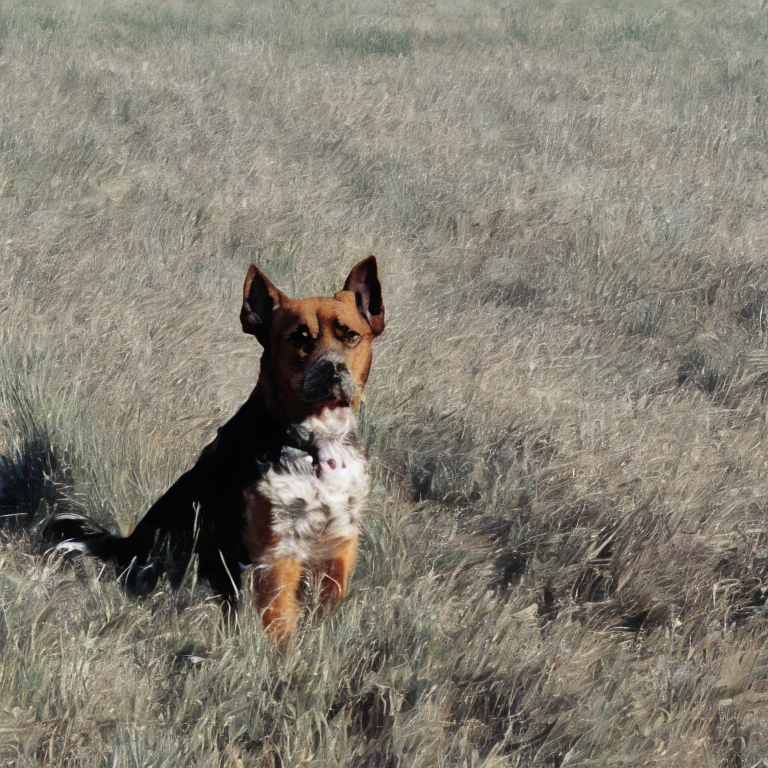}
    \end{subfigure}

    \vspace{0.25em} % space between rows

    % Row 2
    \caption*{Prompt: A treehouse at night}
    \begin{subfigure}{0.45\linewidth}
        \centering
        \includegraphics[width=\linewidth]{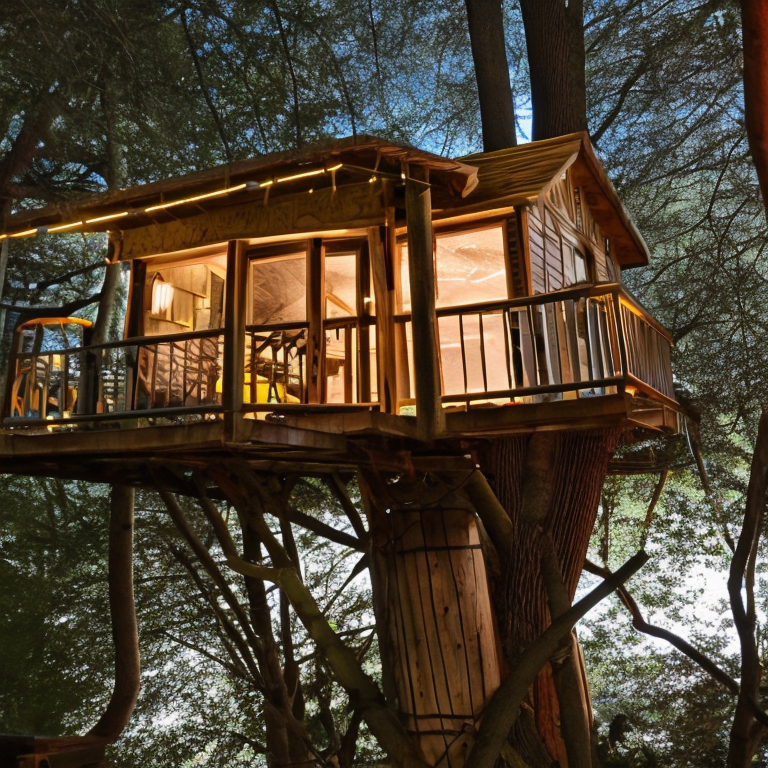}
    \end{subfigure}
    \hfill
    \begin{subfigure}{0.45\linewidth}
        \centering
        \includegraphics[width=\linewidth]{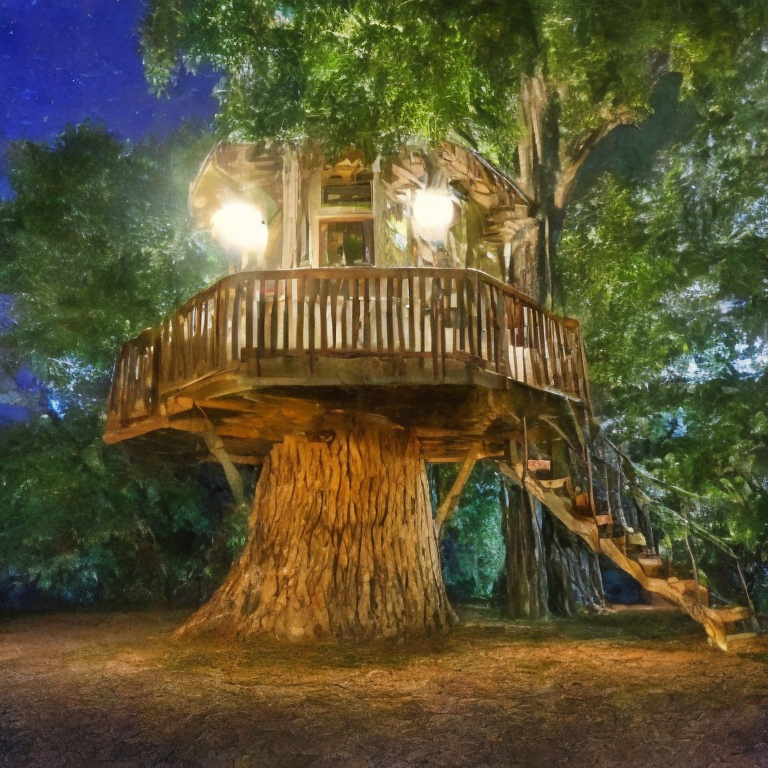}
    \end{subfigure}

    \vspace{0.25em} % space between rows

    % Row 3
    \caption*{Prompt: An ocean sunset}
    \begin{subfigure}{0.45\linewidth}
        \centering
        \includegraphics[width=\linewidth]{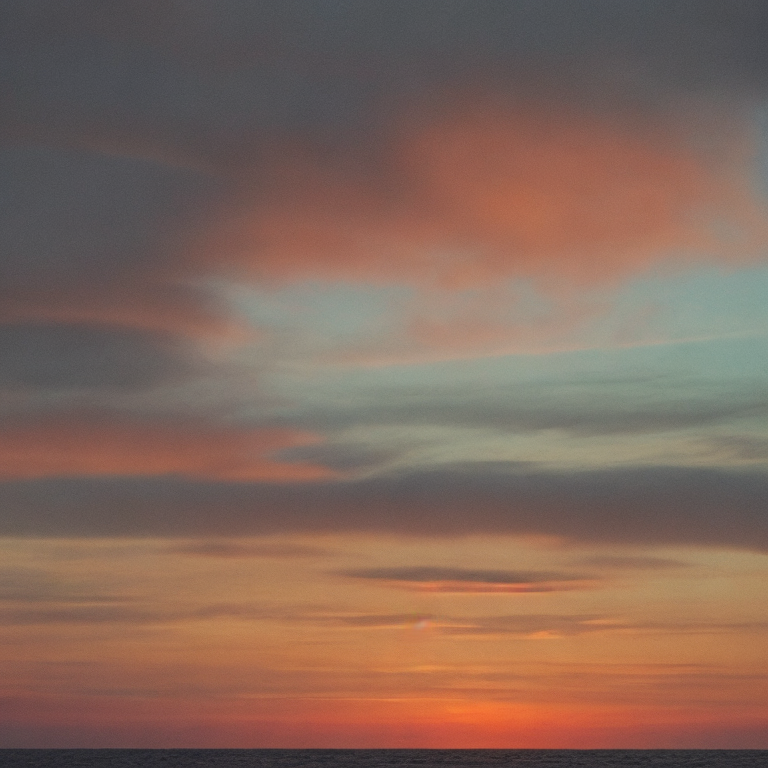}
    \end{subfigure}
    \hfill
    \begin{subfigure}{0.45\linewidth}
        \centering
        \includegraphics[width=\linewidth]{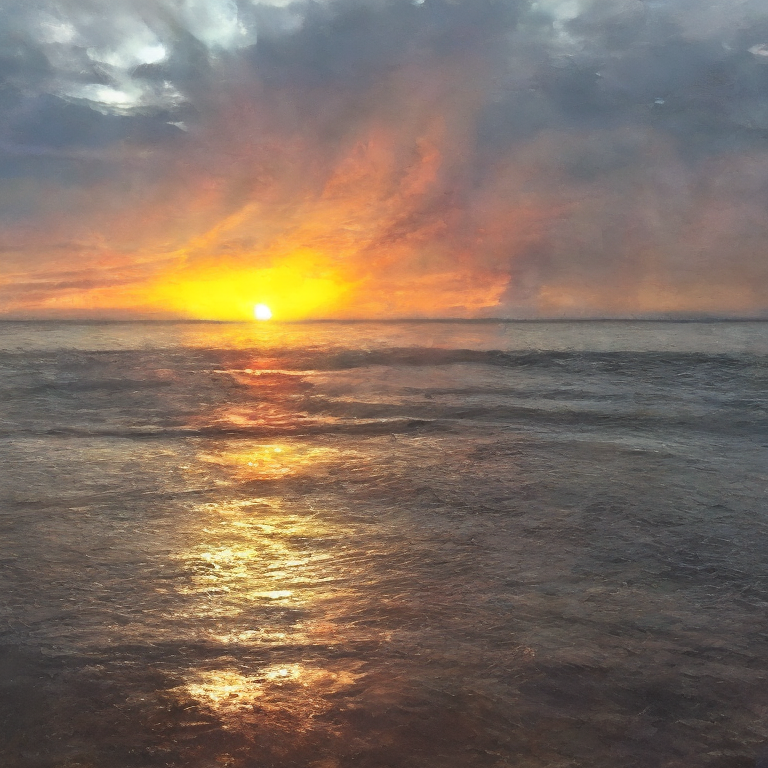}
    \end{subfigure}

    \vspace{0.25em} % space between rows

    % Row 4
    \caption*{Prompt: The Eiffel Tower}
    \begin{subfigure}{0.45\linewidth}
        \centering
        \includegraphics[width=\linewidth]{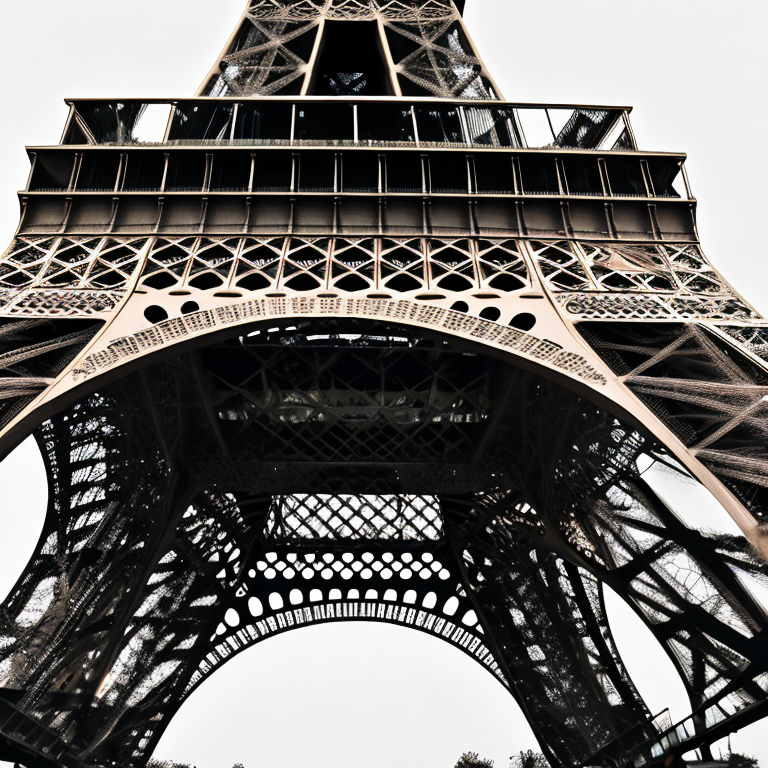}
    \end{subfigure}
    \hfill
    \begin{subfigure}{0.45\linewidth}
        \centering
        \includegraphics[width=\linewidth]{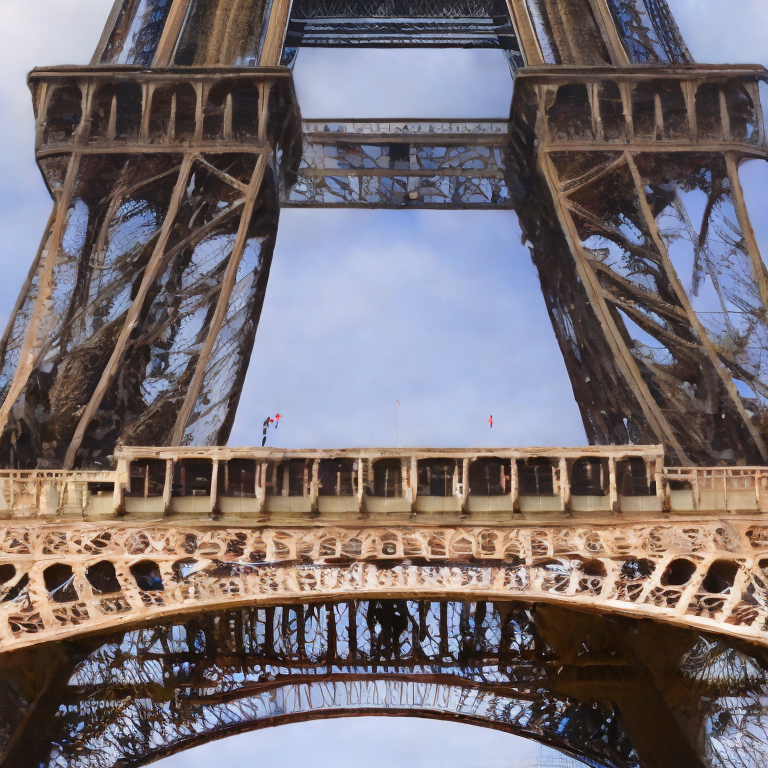}
    \end{subfigure}

    \caption{Base Model vs Optimally Pruned Model at 38.5\% Sparsity}
    \label{fig:base_vs_optimal_config}
\end{figure}

Specifically, Stable Diffusion 2 can be pruned to 38.5\% sparsity by adjusting the text encoder to 47.5\% sparsity and the diffusion generator to 35\% sparsity. This configuration results in minimal loss of image quality while providing a substantial reduction in model size. A comparison of the generated image quality for the baseline and optimal configurations is illustrated in Figure \ref{fig:base_vs_optimal_config}.

\section{Conclusions}

This research presents the first comprehensive study on post-training pruning of text-to-image models, specifically focusing on Stable Diffusion 2. Our findings reveal unexpected behaviors and provide crucial insights into the compression of these complex models.
We demonstrate that Stable Diffusion 2 can be effectively pruned to 38.5\% sparsity with minimal quality loss, achieving a significant reduction in model size. The optimal pruning configuration, with 47.5\% sparsity in the text encoder and 35\% in the diffusion generator, maintains image generation quality while substantially reducing computational requirements. This achievement paves the way for the deployment of advanced text-to-image models on resource-constrained devices, potentially broadening accessibility to this technology.
Contrary to established trends in language model pruning, we found that simple magnitude pruning outperforms more advanced techniques in this context. Additionally, we uncovered the existence of sharp performance thresholds, suggesting a distinctive information encoding mechanism in text-to-image models. These unexpected results highlight the unique challenges and opportunities in pruning text-to-image models, emphasizing the need for specialized approaches in this domain.

Our work opens up new avenues for future research, including the development of pruning techniques tailored to text-to-image models, further investigation into the information encoding mechanisms of these models, and exploration of potential applications in bias identification and model interpretability.

In conclusion, this study not only provides practical insights for the efficient deployment of text-to-image models but also lays the groundwork for a new subfield at the intersection of model compression and text-to-image generation. As these models continue to grow in complexity and capability, the insights and methods presented here will be crucial in ensuring their widespread accessibility and efficient implementation.

\section{Future Work}

% Our experiments and analysis lead to a few future directions. First, exploring post-training pruning techniques that are more well-suited to pruning the text encoder portions of text-to-image models. Magnitude pruning is typically the baseline upon which other pruning techniques try to improve, so the fact that magnitude pruning performs best out of the tested methods is a strong indication that better techniques for pruning the text encoder may exist. For the image diffusion generator, only magnitude pruning was tested in this work. Better methods for post-training pruning of diffusion models exist but could not be tested due to time constraints for this research project. Testing how these methods work in text-to-image models in conjunction with pruning the text encoder is another avenue for future research.  

Our results and analyses suggest several future research directions. %First, there is a need to develop a deeper understanding of which weights and activations are most critical in text-to-image models. 
Our findings indicate that both Wanda and SparseGPT do not effectively extend to these models, while magnitude pruning—typically regarded as the baseline method for pruning—emerges as the best-performing approach in our study. This highlights a significant opportunity for further improvement and optimization in the field. Additionally, our work could be applied to more recent models, such as Stable Diffusion 3 or larger text-to-image models, to assess the generalizability of our findings.

\appendix

\section{Other Results}

\begin{table}[h!]
    \centering
    % \begin{subtable}[t]{0.48\textwidth}
        \centering
        \begin{tabular}{c|ccc}
            \hline
            Total Sparsity & 75:25 & 50:50 & \cellcolor{green!25} 25:75 \\
            \hline
            20\% & 20.92 & 20.9 & \cellcolor{green!25} 18.18 \\
            30\% & 290.97 & 20.81 & \cellcolor{green!25} 18.51 \\
            40\% & 0 & 250.37 & \cellcolor{green!25} 20.48 \\
            50\% & 0 & 226.44 & \cellcolor{green!25} 67.93 \\
            60\% & 0 & 0 & \cellcolor{green!25} 370.47 \\
            \hline

        \end{tabular}
        \caption{FID for Full Model Pruning with OWL}
        \label{tab:owl_full_model_fid_scores}
    % \end{subtable}
\end{table}
    % \hfill
\begin{table}[h!]
% \begin{subtable}[t]{0.48\textwidth}
    \centering
    \begin{tabular}{c|ccc}
        \hline
        Total Sparsity & 75:25 & 50:50 & \cellcolor{green!25} 25:75 \\
        \hline
        20\% & 0.307 & 0.314 & \cellcolor{green!25} 0.317 \\
        30\% & 0.172 & 0.306 & \cellcolor{green!25} 0.316 \\
        40\% & 0 & 0.183 & \cellcolor{green!25} 0.307 \\
        50\% & 0 & 0.182 & \cellcolor{green!25} 0.269 \\
        60\% & 0 & 0 & \cellcolor{green!25} 0.202 \\
        \hline

    \end{tabular}
    \caption{CLIP Score for Full Model Pruning with OWL}
    \label{tab:owl_full_model_clip_scores}
% \end{subtable}
% \caption{FID and CLIP Score for splitting sparsity at different ratios}
% \label{tab:owl_full_model_combined_fid_clip_scores}
\end{table}

\begin{table}[h!]
    \centering
    \scalebox{0.7}{
    \begin{tabular}{cccccc}
        \hline
        Total Sparsity & Text:Image Ratio & Text Sparsity & Image Sparsity & FID & CLIP Score\\
        \hline
        20\% & 75:25 & 53\% & 7\% & 20.92 & 0.307\\
        20\% & 50:50 & 35\% & 14\% & 20.9 & 0.314\\
        20\% & 25:75 & 18\% & 21\% & 18.18 & 0.317\\
        \hline
        30\% & 75:25 & \cellcolor{red!25} 80\% & 10\% & \cellcolor{red!25} 290.97 & \cellcolor{red!25} 0.172\\
        30\% & 50:50 & 53\% & 21\% & 20.81 & 0.306\\
        30\% & 25:75 & 27\% & 31\% & 18.51 & 0.316\\
        \hline
        40\% & 50:50 & \cellcolor{red!25} 71\% & 28\% & \cellcolor{red!25} 250.37 & \cellcolor{red!25} 0.183\\
        40\% & 25:75 & 35\% & 42\% & 20.48 & 0.307\\
        \hline
        50\% & 50:50 & \cellcolor{red!25} 89\% & 35\% & \cellcolor{red!25} 226.44 & \cellcolor{red!25} 0.182\\
        50\% & 25:75 & 44\% & 52\% & 67.93 & 0.269\\
        \hline
        60\% & 25:75 & 53\% & 63\% & 370.47 & 0.202\\
        \hline

    \end{tabular}
    }
    \caption{Model Performance greatly suffers when component sparsities violate the identified drop-off thresholds}
    \label{tab:owl_full_model_sparsities_results}
\end{table}

\begin{figure}[h!]
    \centering
    \subfloat[30\%]{
        \includegraphics[width=0.225\textwidth]{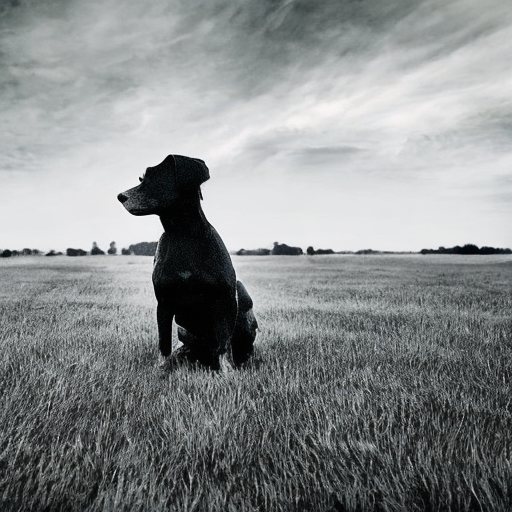}
    }
    \subfloat[40\%]{
        \includegraphics[width=0.225\textwidth]{images/pruning_examples/diffusion_only-0.4.png}
    }
    
    \subfloat[50\%]{
        \includegraphics[width=0.225\textwidth]{images/pruning_examples/diffusion_only-0.5.png}
    }
    \subfloat[60\%]{
        \includegraphics[width=0.225\textwidth]{images/pruning_examples/diffusion_only-0.6.png}
    }
    \caption{Magnitude Pruning Diffusion Generator causes linear deterioration of quality}
    \label{fig:diffusion_magnitude_only_gradual_shift}
\end{figure}

\begin{table}[h!]
    \centering
    \scalebox{0.875}{
    \begin{tabular}{ccccc}
        \hline
        Total Sparsity & Text Sparsity & Image Sparsity & FID & CLIP Score\\   
        \hline
        \cellcolor{gray!40}0\% & \cellcolor{gray!40}0\% & \cellcolor{gray!40}0\% & \cellcolor{gray!40}18.07 & \cellcolor{gray!40}0.314\\
        53.5\% & 62.5\% & 50.0\% & 241.71 & 0.193\\
        51.0\% & 60.0\% & 47.5\% & 122.76 & 0.236\\
        48.5\% & 57.5\% & 45.0\% & 59.12 & 0.273\\
        46.0\% & 55.0\% & 42.5\% & 33.09 & 0.294\\
        43.5\% & 52.5\% & 40.0\% & 24.96 & 0.304\\
        41.0\% & 50.0\% & 37.5\% & 20.75 & 0.309\\
        \cellcolor{green!25}38.5\% & \cellcolor{green!25}47.5\% & \cellcolor{green!25}35.0\% & \cellcolor{green!25}17.97 & \cellcolor{green!25}0.311\\
        36.0\% & 45.0\% & 32.5\% & 16.79 & 0.314\\
        33.5\% & 42.5\% & 30.0\% & 16.45 & 0.315\\
        \hline

    \end{tabular}
    }
    \caption{Pruning Stable Diffusion 2 to found thresholds}
    \label{tab:final_pruning_configs_results_owl_version}
\end{table}

\bibliographystyle{ACM-Reference-Format}
% \bibliography{sample-bibliography} 
\bibliography{main}

\end{document}